\documentclass[journal]{IEEEtran}
\usepackage{graphicx}
\usepackage{textcomp}  
\usepackage{amsfonts}
\usepackage{amsmath}
\usepackage{booktabs}
\usepackage{multirow}
\usepackage{algorithm}
\usepackage{algorithmic}
\usepackage{float}
\usepackage{cite}

\renewcommand{\algorithmicrequire}{ \textbf{Input:}} 
\renewcommand{\algorithmicensure}{ \textbf{Output:}} 

\ifCLASSINFOpdf
\else
\fi

\ifCLASSOPTIONcompsoc
	\usepackage[caption=false,font=normalsize,labelfont=sf,textfont=sf]{subfig}
\else
  \usepackage[caption=false,font=footnotesize]{subfig}
\fi

\begin{document}
%
\title{Bi-directional Exponential Angular Triplet Loss \\for RGB-Infrared Person Re-Identification}
%
%
%

\author{Hanrong~Ye,
        Hong~Liu,
        Fanyang~Meng,
        and~Xia~Li

\thanks{H. Ye, H. Liu and X. Li are with Key Laboratory of Machine Perception, Shenzhen Graduate School, Peking University, Beijing 100871, China. E-mail: \{leoyhr, hongliu, ethanlee\}@pku.edu.cn. F. Meng is with Pengcheng Laboratory, Shenzhen 518000, China. E-mail: mengfy@pcl.ac.cn.
	This work 	was supported  by National Natural Science Foundation of China (U1613209), National Natural Science Foundation of Shenzhen (No. JCYJ20190808182209321), The Verification Platform of Multi-tier Coverage Communication Network for Oceans (LZC0020)  and Youth Program of National Natural Science Foundation of China (61906103). Codes available at https://github.com/prismformore/expAT.
 Corresponding author: Hong Liu.  }
}

\markboth{Journal of \LaTeX\ Class Files}
{Ye \MakeLowercase{\textit{et al.}}: Bi-directional Exponential Angular Triplet Loss for RGB-Infrared Person Re-Identification}

\maketitle

\begin{abstract}
RGB-Infrared person re-identification (RGB-IR Re-ID) is a cross-modality matching problem, where the modality discrepancy is a big challenge.
Most existing works use Euclidean metric based constraints to resolve the discrepancy between features of images from different modalities.
However,  these methods are incapable of learning angularly discriminative feature embedding because Euclidean distance cannot measure the included angle between embedding vectors effectively.
As an angularly discriminative feature space is important for classifying the human images based on their embedding vectors,
in this paper, we propose a novel ranking loss function, named Bi-directional Exponential Angular Triplet Loss, to help learn an angularly separable common feature space by explicitly constraining the included angles between embedding vectors.
Moreover, to help stabilize and learn the magnitudes of embedding vectors, we adopt a common space batch normalization layer.
The quantitative and qualitative experiments on the SYSU-MM01 and RegDB dataset support our analysis.
On SYSU-MM01 dataset, the performance is improved from 7.40\%~/~11.46\% to 38.57\%~/~38.61\%  for rank-1 accuracy~/~mAP compared with the baseline.
The proposed method can be generalized to the task of single-modality Re-ID and improves the rank-1 accuracy~/~mAP from 92.0\%~/~81.7\% to 94.7\%~/~86.6\% on the Market-1501 dataset, from 82.6\%~/~70.6\% to 87.6\%~/~77.1\% on the DukeMTMC-reID dataset.
\end{abstract}


\IEEEpeerreviewmaketitle

\section{Introduction}
\label{sec:introduction}
\IEEEPARstart{P}{erson} re-identification~(Re-ID) aims at recognizing the same person across cameras of non-overlapping spaces. Given an image of a person captured from a camera~(query set),  the algorithm is expected to retrieve the images of the same person from the images captured under different views~(gallery set).
It has a wide range of applications in video analysis, intelligent surveillance, and other systems~\cite{ccloy09,xwang13,kissme,zheng2015towards,ma2015body,you2016top,yang2014salient}.

Re-ID is a challenging task influenced by lots of complex factors like occlusion, intra-class variations  as well as inter-class similarity.
Intra-class variations mean that the same person may look different in different views. Typical intra-class variations include various illumination conditions, human poses, change of clothes and resolution etc.
Inter-class similarity means that different people may look similar in different views, especially when they put on similar outfits.

Many related works have been proposed in recent years~\cite{sgg2016highly,kniaz2018thermalgan,li2018harmonious,yu2018tpami,weishiicip2018,li2015multi,deng2018image}, most focus on Re-ID among RGB cameras~(RGB Re-ID).
However, as RGB cameras fail to work in the dark environment~(indoor or nighttime), it is infeasible to conduct Re-ID with only RGB cameras in the real-world scenarios. To deal with this problem, the infrared~(IR) cameras, which are able to capture visual information with IR light, have been commonly used in surveillance systems nowadays.
The use of IR cameras provides rich information for Re-ID in the dark, and enables the research in Re-ID under bad lighting condition.

Different from RGB cameras, which are able to decouple visible light into three channels~(red, green and blue) according to wavelength, surveillance IR cameras output a single-channel response map of some certain infrared wavelength interval based on the characteristics curve. Therefore, the data distributions of these two modalities are discrepant, as shown in Figure~\ref{img_res_vis_sysu}.
This phenomenon is called ``modality discrepancy''~\cite{tone,GANCVPR19}.

\begin{figure}[t]
	\centering
	\includegraphics[scale=0.28]{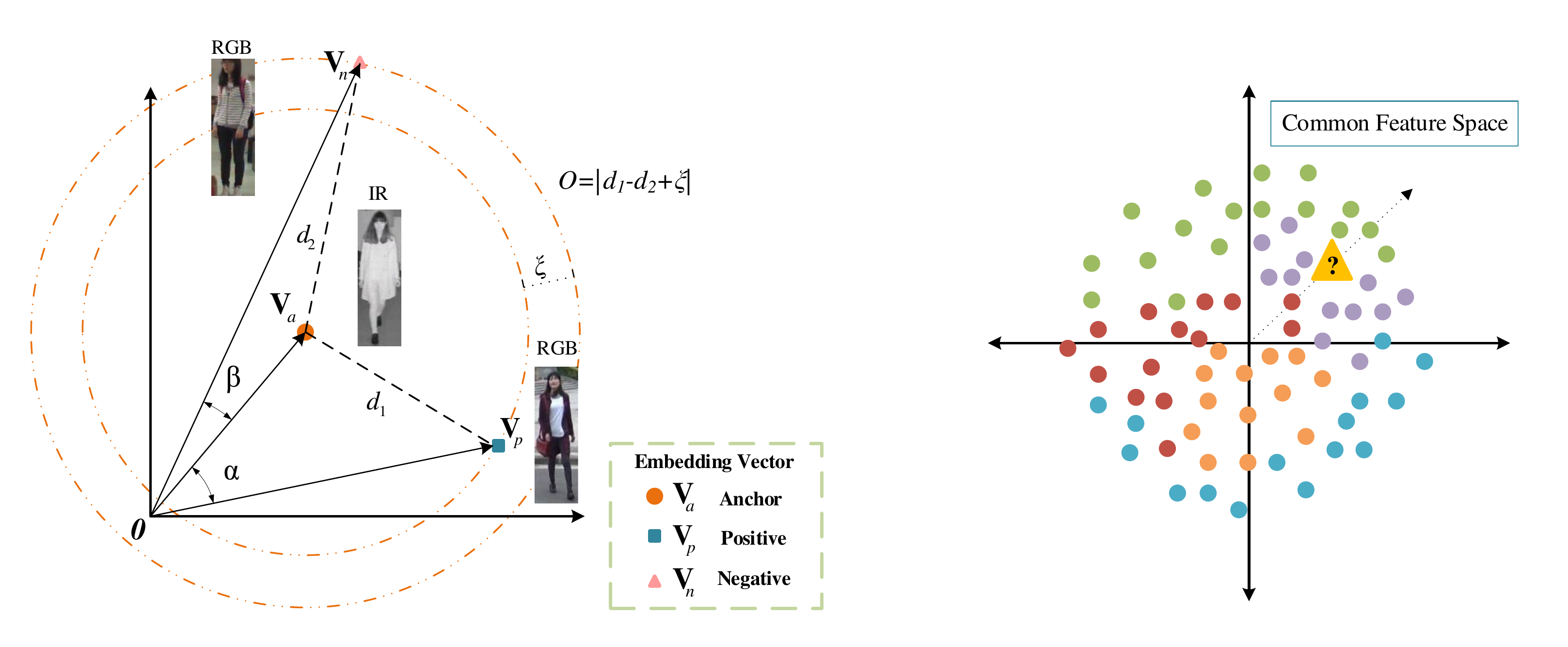}
	\caption{Illustration of the issue in triplet loss. \textbf{Left:} In the common feature space, suppose that $\mathbf{V}_a$, $\mathbf{V}_p$ and $\mathbf{V}_n$ are embedding vectors of anchor image, positive image and negative image respectively.
	When the objective function $O$ converges~($O \rightarrow 0$),  the embedding vectors of different classes may share more similar orientations than those of the same class. Thus the triplet loss does not promise an angularly separable feature space for different classes. 
	\textbf{Right:} Diagram of the common feature space after training with triplet loss. Circles with different colors represent features from different classes. Since the features are mixed angularly in the common feature space,  it is hard to correctly  tell which class it belongs to for an embedding vector obtained from the testing dataset (denoted as a yellow triangle in the diagram).   Best viewed in color.
	}
	\label{img_at}
\end{figure}

To bridge the modality gap of data distribution between RGB and IR images, recent works~\cite{VT,cmgan,GANCVPR19} use Euclidean metric constraints like Triplet Loss~\cite{tripletloss,defensetriplet2017,chen2017beyond,Liuacmmm} to encourage the Euclidean distances between features of the same label to be smaller than those between different labels.
Although these methods enjoy success by learning the distances between features explicitly, there is an inherent inadequacy in the design of Euclidean metric based loss function:
triplet loss cannot constrain the included angles between embedding vectors effectively and thus fails to separate the orientations of embedding vectors in the common space.
As illustrated in the Figure~\ref{img_at}, the included angles ($\alpha$ and $\beta$) of positive pairs and negative pairs are uncertain when the triplet loss converges.
The positive  pairs may even have larger included angles than negative pairs.
This phenomenon makes the feature generator unable to separate the feature embedding vectors of different classes in the common space angularly when inferring, as shown intuitively in Figure~\ref{img_at} and Figure~\ref{img_pca}.

What's more, in the training phase, the final linear layer (without bias term) for classification loss benefits from an angularly more discriminative common feature space.
The linear layer is actually computing dot product between feature vectors $\{\vec{a}_i, i\in[1,N]\}$ and weight vectors $\{\vec{b}_i, i\in[1,C]\}$ of different classes.
As $\vec{a}_i  \vec{b}_j = |\vec{a}_i||\vec{b}_j|cos(\theta(i,j))$, for each feature vector  $\vec{a}_i$,  the ranking of $\{\vec{a}_i  \vec{b}_j, j\in[1,C]\}$ depends on the included angles $\{\theta(i,j),  j\in[1,C]\}$ and the magnitudes of weight vectors $\{|\vec{b}_j|, j\in[1,C]\}$.
Since the magnitudes of the weight vectors $\{|\vec{b}_i|, i\in[1,C]\}$ should be similar to each other (otherwise the linear classifier will have strong prior bias towards some  particular classes),
the ranking is dominated by the included angles, or the directions, of the feature vectors.
An angularly separable common feature space helps the classification loss in this way.

To address the issue of Euclidean metric based triplet loss, we put forward a cross-modality ranking loss focusing on the included angles between embedding vectors  generated from different domains. The proposed constraint, called Bi-directional Exponential Angular Triplet (expAT) Loss, utilizes cosine distance  as a direct measure of included angles.
What's more, since cosine distance cannot restrict the magnitude of embedding vectors, a special Common Space Batch Normalization layer is designed to assist Bi-directional Exponential Angular Triplet Loss in stabilizing and learning the magnitudes of embedding vectors. With these components, we design an end-to-end single-stream framework for RGB-IR Re-ID.

The main contributions of this paper are summarized as follows:
\begin{itemize}
	\item A novel included angle based cross-modality ranking constraint~(expAT Loss) is proposed to address the difficulty in learning angularly discriminative feature embedding. The proposed loss is distinguished from the existing methods and easy to implement. Feature visualization manifests that our method helps learn an intuitively more separable common feature space compared with triplet loss.
	
	\item  To help with stabilizing the magnitudes of embedding vectors, we adopt a variant of batch normalization layer on the common feature space, named ``Common Space Batch Normalization'', which brings large performance improvement working with the proposed included angle based metric learning loss function.

	\item An end-to-end single-stream framework is proposed for the challenging  RGB-IR Re-ID task. The proposed method can be generalized to the cross-modality RGB-thermal Re-ID and single-modality RGB Re-ID task. Experiment results on the large-scale SYSU-MM01 dataset, RegDB dataset, Market-1501 dataset and DukeMTMC-reID dataset demonstrate significant improvement over the baseline, and our proposal achieves the state-of-the-art performance.

\end{itemize}

\begin{figure}
	\centering
	\includegraphics[scale=0.12]{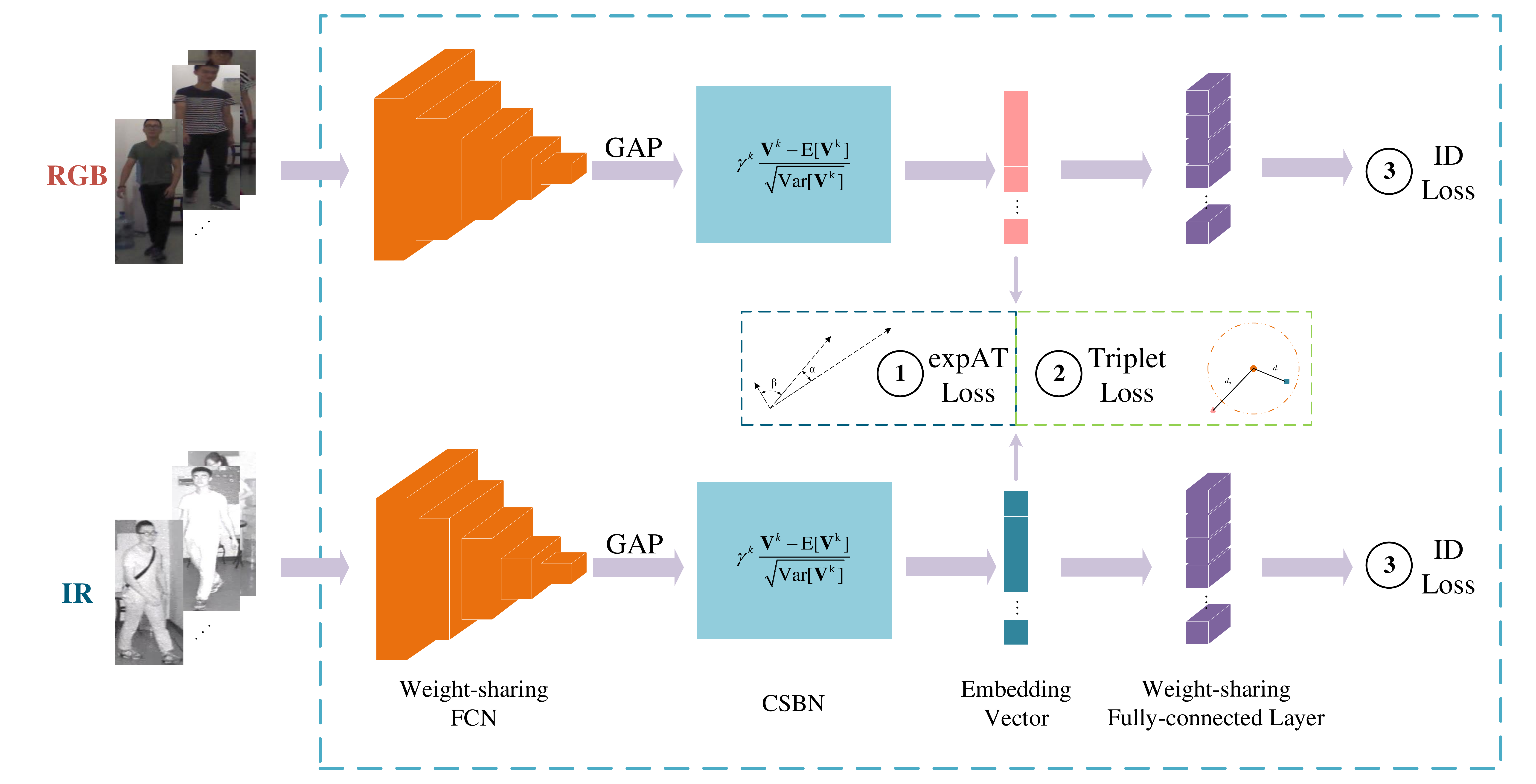}
	\caption{Illustration of the proposed single-stream framework. RGB and IR images are processed by a Fully Convolutional Network~(FCN) to extract semantic feature maps. The feature maps are down-sampled by the Global Average Pooling~(GAP) layer, and then adjusted by the Common Space Batch Normalization~(CSBN) layer to obtain the final embedding vectors. A fully-connected layer is adopted to classify the identity labels of input images. The FCN, CSBN and fully-connected layer are weight-sharing. There are three constraints available for experiments: { \textcircled{\tiny{1}}} The proposed Bi-directional Exponential Angular Triplet~(expAT) Loss; \textcircled{\tiny{2}} Triplet Loss; \textcircled{\tiny{3}} Identity~(ID) Loss.
	}
	\label{img_at_framework}
\end{figure}

\section{Related Work}
In this section, we introduce some exciting works related to the topics of RGB Re-ID and RGB-IR Re-ID.

\subsection{RGB Re-ID}
Popular RGB Re-ID methods~\cite{zhengreview2016} work on generating the discriminative semantic representations for the person based on appearance with low-level features (such as color, shape and texture) or high-level features (such as attributes and deep features). With the development of large-scale parallel computing, deep learning based methods show more promising performance in these years.

To obtain discriminative features, the supervised deep learning based Re-ID methods can be categorized into two classes:  representation learning and metric learning.
In the representation learning methods, Re-ID is viewed as a classification problem and the embedding vectors before fully-connected layer are used for similarity matching.
These embedding vectors are expected to describe the unseen human images well after training on the labeled human images.
Recent works include part based~\cite{attributes2012,pcbrpp2018,zheng2015partial,feris2014attribute}, attention based~\cite{wang2018mancs,li2018harmonious,liu2017end,xu2018attention}, GAN based methods~\cite{zhong2019camstyle,qian2018pose,zheng2017unlabeled,wei2018person}, etc.
To generate more discriminative features, the metric learning methods are introduced in this field to mine the  association of data points.
In the metric learning methods, discriminative features are learned by comparing feature distances between different samples. These methods~\cite{BDB,MGN,weishirdc2013,cheng2016person,liu2018discriminatively} pull the features of the same identity together and push those from different identities away with specially designed loss functions like contrastive loss and triplet loss.

\textbf{Angular constraints}. An angle based loss function has been studied in the field of RGB Re-ID, named cosine softmax loss~\cite{wojke2018}.
Cosine softmax loss aims at improving the angular discrimination of embedding vectors generated by the models.
It develops sophisticated softmax based \textit{classifier loss} with L2 normalization.
Differently, our method works on directly constraining the included angles between embedding vectors based on the \textit{ranking loss}.

\subsection{RGB-IR Re-ID}
Although the infrared (IR) cameras have been widely used to capture visual information in the dark for security need, the research of RGB-IR Re-ID  is rather rare.
Recently, to empower research in this field, researchers  propose the first large-scale RGB-IR Re-ID dataset for surveillance scenario named SYSU-MM01~\cite{SYSUMM01}.
In this benchmark, the images from RGB cameras form gallery set, and those from IR cameras form query set, which resembles the real-world requirement.

Several works have been proposed for RGB-IR Re-ID including both single-stream and two-stream networks.
For the single-stream model, Wu et al.~\cite{SYSUMM01}  propose a deep zero-padding strategy to help networks learn domain-specific nodes. They pad the multi-modality inputs into a domain-specific manner.
Dai et al.~\cite{cmgan} and Wang et al.~\cite{GANCVPR19} design special adversarial training pipelines to guide the learning of cross-modality representations and achieve the state-of-the-art performance.

For the two-stream model, HCML~\cite{tone} and BDTR~\cite{VT} exploit cross-modality constraints with both metric learning and representation learning. MSR~\cite{msr2019tip} develops a two-stream network to extract modality-specific as well as modality-shared representation  with the help of view classification~\cite{feng2018viewreid}.  These works also tackle another cross-modality person re-identification task between RGB cameras and thermal cameras. The thermal cameras collect far-infrared radiation emitted from human body heat~\cite{nguyen2017} and produce response maps that are very different from those of surveillance IR cameras, which emit IR light actively.
For comparison with these algorithms, we also discuss the performance of our methods on RegDB dataset~\cite{nguyen2017}, which is a popular benchmark in this research field.

\section{Proposed Method}
\subsection{Overview}
RGB-IR Re-ID is challenging due to the modality discrepancy. In this work, we focus on minimizing the cross-modality gap and obtaining discriminative features by effectively exploiting the constraints on the common feature space.

In this section, we will introduce the structure of the proposed framework,
which can be divided into two components:  the feature extractor and the common feature space constraints. As shown in the Figure~\ref{img_at_framework}, the pipeline of the proposed method is summarized as follows: A fully convolutional backbone network is adopted as the feature extractor for both RGB and IR domains to extract high-level semantic features from input. Then, these features are mapped to the common feature space.
In the common feature space, the identity loss and the proposed cross-modality ranking loss  work  as constraints to ensure that the discrepancy between RGB and IR embedding vectors is reduced effectively with respect to included angle.

\subsection{Feature extractor}
\label{sec_network}
The role of feature extractor is to map the input from different modalities into the common feature space. The network structure is shown in the Figure~\ref{img_at_framework}.
A popular backbone model, ResNet50~\cite{resnet2016}, is selected for both RGB and IR domains to extract high-level semantic features of input.

To take advantage of knowledge transfer, we pre-train the model on the ImageNet~\cite{imagenet_cvpr09}.  Although IR images contain no color information, ImageNet pre-trained models can still help with capturing their contour and texture information. Specifically, we adopt the fully convolutional network of ImageNet pre-trained ResNet50 removing its final fully-connected layer. To obtain feature maps with a higher spatial resolution for finer processing, we remove the down-sampling operation of the last convolution block. This backbone model is denoted as $\mathcal{F}$.

Global average pooling layer ($\mathcal{G}$) is employed on the top of feature extractor to acquire a compact embedding vector in the common space. Then, a special common space batch normalization layer ($\mathrm{CSBN}$) is adopted here and will be explained in details at the following illustration of common feature space constraints.

Precisely, given an input image $I$, the extracted embedding vector $\mathbf{V}$ can be calculated as:
\begin{equation}
\mathbf{V} = \mathrm{CSBN}(\mathcal{G}(\mathcal{F}(I))).
\label{equ_extractor}
\end{equation}

In this work, we propose to use a single-stream framework to process both the RGB and IR images, i.e. the feature extractors of the RGB and IR inputs share weights.
Notably, although other works~\cite{VT} use a fully-connected embedding layer with activation function to further process the embedding vectors from different modalities, we observe that  redundant non-linear layer leads to worse performance as shown in the experiment section. The backbone network alone is powerful enough to learn feature embedding directly.

\subsection{Common Feature Space Constraints}
\label{sec_constraints}
In this part,  we will firstly introduce the cross-modality version of the  widely used triplet loss, analyze its issue, and then deduce a novel cross-modality cosine ranking loss. Later, we would illustrate the proposed common space batch normalization.

\subsubsection{\textbf{Bi-directional Triplet Loss} }

\noindent The motivation of triplet loss~\cite{defensetriplet2017}  is to obtain ``good'' embedding vectors, i.e., to make the distances between embedding vectors of the same class smaller than those between different classes.

Triplet loss makes sure that given an anchor image $I_a$, the
embedding vector of a positive image $I_p$ belonging to the same class
(which means the same person in this task) is closer to the anchor's embedding vector than that of a negative image  $I_n$ belonging to another class, by a
margin $\xi > 0$. The objective of this constraint is that eventually, all embedding vectors of the same class will be closer to each other than to any from different classes. This criterion is formulated as:
\begin{equation}
O=|d_1 - d_2 + \xi| \rightarrow 0.
\label{equ_objective}
\end{equation}

Following this idea and denoting the Euclidean distance between $\mathbf{X}$ and $\mathbf{Y}$  as $\mathbb{D}(\mathbf{X}, \mathbf{Y})$, the triplet loss is written as:

\begin{equation}
\mathcal{L'}_{triplet} = \frac{1}{N}  \!\!\!\!\!\! \sum_{(I_a, I_p, I_n)} [\mathbb{D}(\mathbf{V}_a, \mathbf{V}_p) - \mathbb{D}(\mathbf{V}_a, \mathbf{V}_n) + \xi]_+,
\label{equ_triplet}
\end{equation}
here, $N$ is the mini-batch size, $\mathbf{V}_a$, $\mathbf{V}_p$ and $\mathbf{V}_n$ are embedding vectors of anchor image, positive image and negative image separately in an image triplet $(I_a, I_p, I_n)$. The clamping function~$[x]_+$ is:
\begin{equation}
[x]_+ = \max(x, 0).
\end{equation}

\noindent\textbf{Bi-directional manner.}
Different from the single-modality problem, as the matching is conducted across two modalities, it is more important to connect embedding vectors from different domains and \textit{resolve the discrepancy between different modalities}. Thus, when the anchor term is sampled from RGB domain, the positive and negative terms are sampled from IR domain, and vice versa. This manner is named ``bi-directional''~\cite{VT}.

The  cross-modality version of triplet loss  $\mathcal{L}_{triplet}$ includes RGB anchor based term $\mathcal{L}_{triplet}^{rgb}$ and IR anchor based  term $\mathcal{L}_{triplet}^{ir}$. Then we have the bi-directional triplet loss:
\begin{equation}
\begin{aligned}
\mathcal{L}_{triplet} &= \mathcal{L}_{triplet}^{rgb} + \mathcal{L}_{triplet}^{ir} \\
&\!\!\!\!\!\!\!\!\! =  \frac{1}{N}  \!\!\!\!\!\! \sum_{(I_a^{rgb}, I_p^{ir}, I_n^{ir})} [\mathbb{D}(\mathbf{V}_a^{rgb}, \mathbf{V}_p^{ir}) - \mathbb{D}(\mathbf{V}_a^{rgb}, \mathbf{V}_n^{ir}) + \xi]_+ \\
&\!\!\!\!\!\!\!\!\! +  \frac{1}{N} \!\!\!\!\!\! \sum_{(I_a^{ir}, I_p^{rgb}, I_n^{rgb})} \![\mathbb{D}(\mathbf{V}_a^{ir}, \mathbf{V}_p^{rgb}) - \mathbb{D}(\mathbf{V}_a^{ir}, \mathbf{V}_n^{rgb}) + \xi]_+ .
\end{aligned}
\label{equ_cmtloss}
\end{equation}

In triplet $(I_a^{rgb}, I_p^{ir}, I_n^{ir})$, $I_a^{rgb}$ and $I_p^{ir}$ share the same label while $I_a^{rgb}$ and $I_n^{ir}$ have different labels; in triplet $(I_a^{ir}, I_p^{rgb}, I_n^{rgb})$, $I_a^{ir}$ and $I_p^{rgb}$ share the same label while $I_a^{ir}$ and $I_n^{rgb}$ have different labels.

\begin{figure*}[h]
	\centering
	\subfloat[margin=50pt][Triplet Loss]{ \includegraphics[width=0.48\linewidth,trim=0 30 0 0,clip]{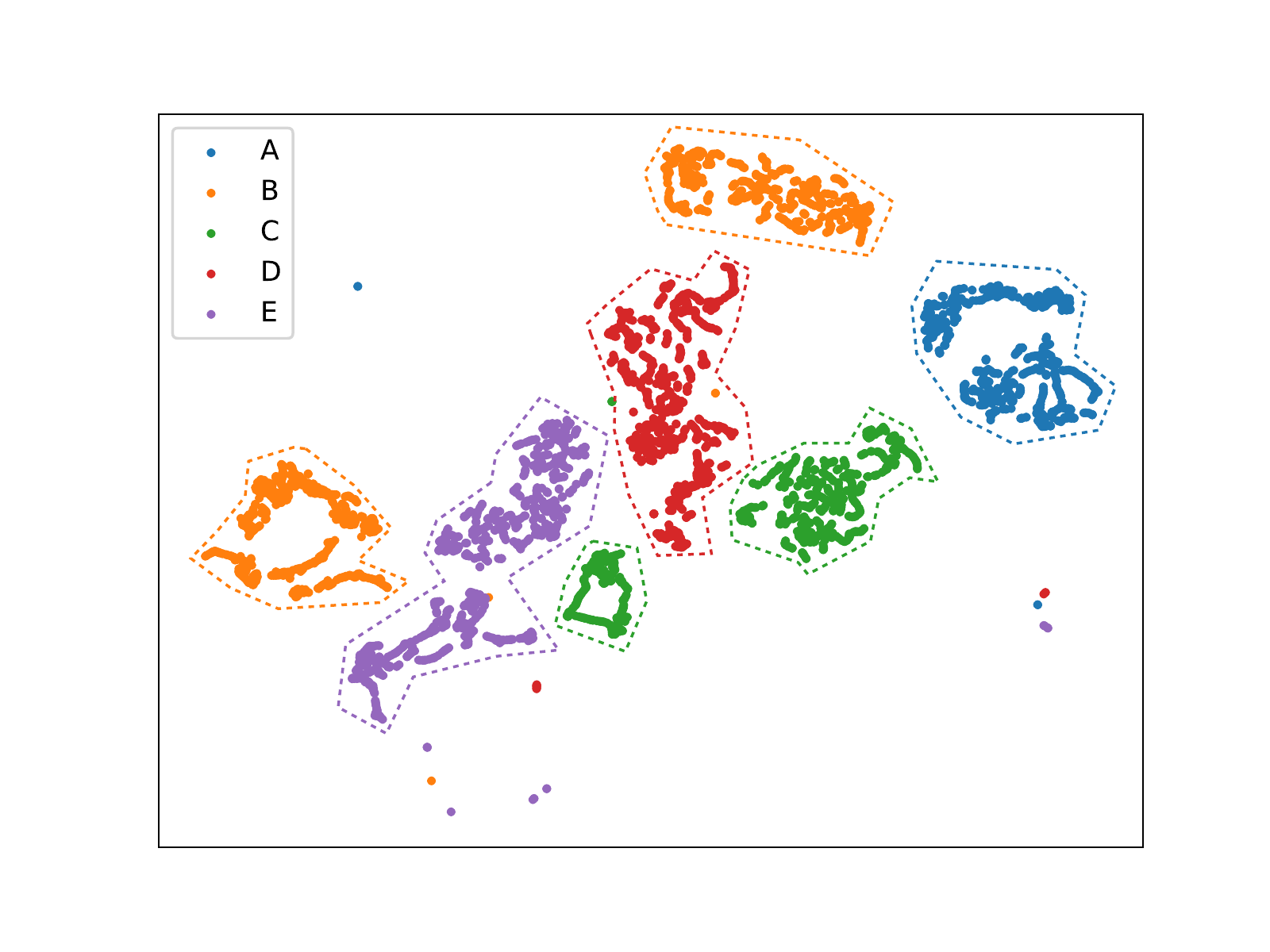}
	}
  \hfill
	\subfloat[expAT Loss]{
\includegraphics[width=0.48\linewidth,trim=0 30 0 0,clip]{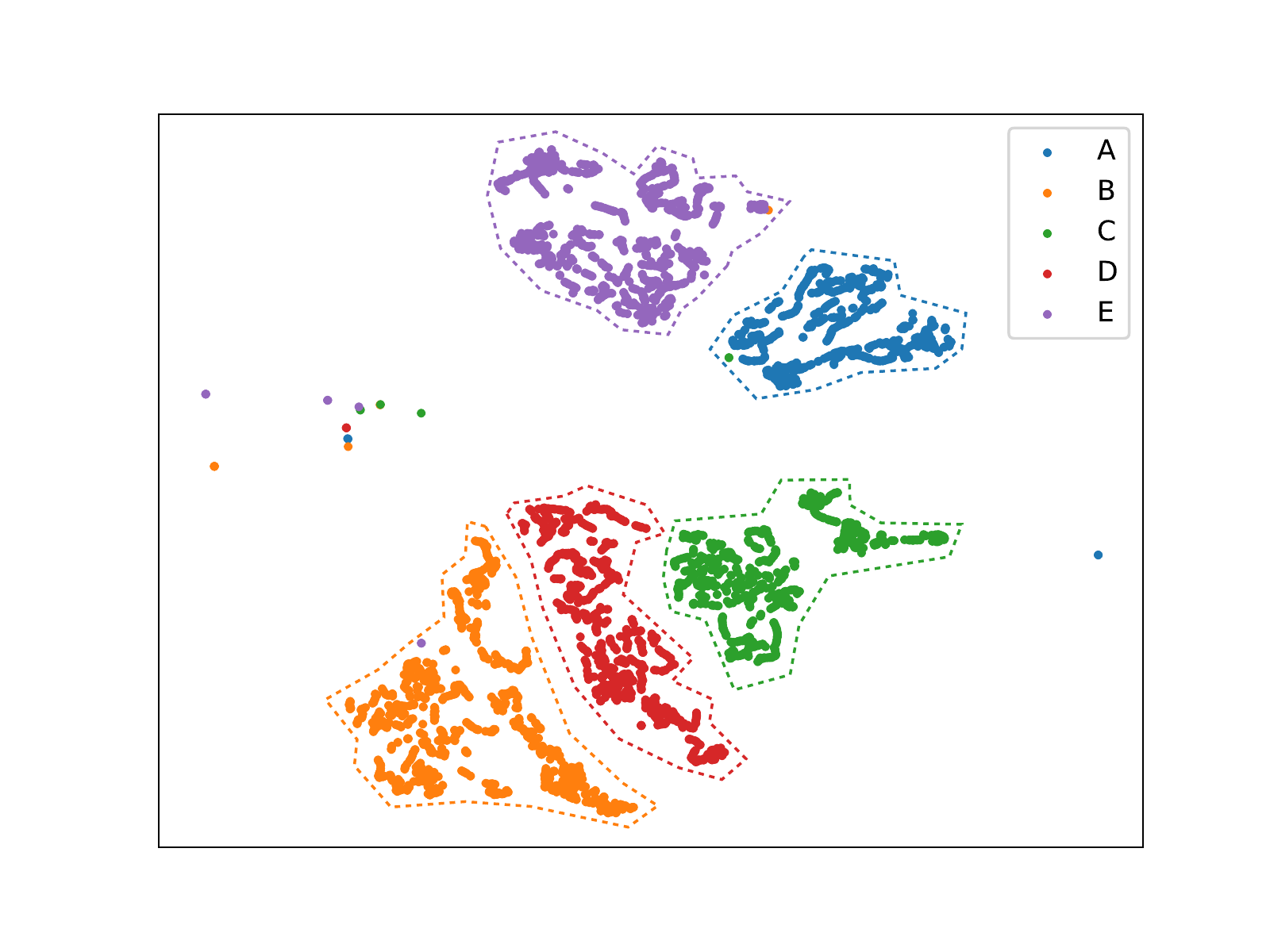}
	}
	\caption{2-D visualization of the common feature space on the testing set
of SYSU-MM01 dataset.  With expAT loss, the embedding
vectors of different classes are more separable in feature space. ``A''-``E'' represent different labels.
 (Best viewed in color.)}
	\label{img_pca}
\end{figure*}

\subsubsection{\textbf{Issue in Triplet Loss}}

\noindent As illustrated in the  Figure~\ref{img_at}, when the constraint function (Equation \ref{equ_objective}) satisfied (which means $\mathbb{D}(\mathbf{V}_a,\mathbf{V}_n) - \mathbb{D}(\mathbf{V}_a,\mathbf{V}_p)\rightarrow \xi, \xi>0$),
the included angles between embedding vectors are uncertain and the embedding vectors from different classes may share more similar directions than those from the same class ($\beta<\alpha$).
This phenomenon makes the embedding vectors of different classes ``mixed'' together in common space and fail to cluster, especially for the unseen samples from the testing dataset. 2-D visualization of the common feature space with triplet loss is presented in the Figure~\ref{img_pca}. On the contrary, a separable common feature space with embedding vectors from different classes located in isolated areas is desirable.

\subsubsection{\textbf{Bi-directional Exponential Angular Triplet Loss} }

\noindent As the triplet loss has difficulty in separating the orientations of embedding vectors, to address this issue, we design a new cross-modality ranking loss called Bi-directional Exponential Angular Triplet~(expAT) Loss. Instead of Euclidean distance, expAT loss utilizes cosine distance function to measure the difference between embedding vectors.

As the cosine distance $\mathbb{C}$ only pays attention to the included angle between vectors, it puts a strong constraint on embedding vectors to learn the proper orientations in common space.

Firstly, we begin with a naive idea of cosine distance based triplet loss is:
\begin{equation}
\begin{aligned}
\mathcal{L}_{cos}  &=  \frac{1}{N}  \!\!\!\!\!\!  \sum_{(I_a, I_p, I_n)} \!\!\!\!\!\! [\mathbb{C}(\mathbf{V}_a, \mathbf{V}_p) -  \mathbb{C}(\mathbf{V}_a, \mathbf{V}_n) + \xi]_+, \\
\end{aligned}
\end{equation}
here, cosine distance $\mathbb{C}(\mathbf{X},\mathbf{Y}) = 1 - \cos(\mathbf{X},\mathbf{Y}) =
1 - \frac{\mathbf{X}\cdot \mathbf{Y}}{|\mathbf{X}||\mathbf{Y}|}$.  Then we have:
\begin{equation}
\begin{aligned}
\mathcal{L}_{cos}  &= \frac{1}{N}  \!\!\!\!\!\! \sum_{(I_a, I_p, I_n)} \!\!\!\!\!\! [ (1 - \cos(\mathbf{V}_a,\mathbf{V}_p))
- (1 - \cos(\mathbf{V}_a,\mathbf{V}_n))
+ \xi
]_+ \\
&= \frac{1}{N}  \!\!\!\!\!\! \sum_{(I_a, I_p, I_n)} \!\!\!\!\!\! [\cos(\mathbf{V}_a, \mathbf{V}_n) - \cos(\mathbf{V}_a,\mathbf{V}_p)
+ \xi]_+.
\end{aligned}
\end{equation}

There are two flaws in this loss function. First, our goal is to make $\mathbf{V}_a$ and $\mathbf{V}_n$ tend to be \textit{uncorrelated} (orthogonal) in the feature space, which means  $\cos(\mathbf{V}_a, \mathbf{V}_n) \to 0$ instead of $\cos(\mathbf{V}_a, \mathbf{V}_n) \to -1$ (which is negative correlation). Thus a clamping function is adopted to constrain $\cos(\mathbf{V}_a, \mathbf{V}_n) $.
Second, as $\cos(\mathbf{X},\mathbf{Y}) \in [-1,1]$, the overall clamping function is removed with a proper choice of margin $\xi$.
When margin $\xi$  is no less than 1, the function can always be non-negative.
Thus we simply set $xi$ to 1 to keep the hyper-parameters simple.

With these motivations, we heuristically propose a loss function called angular triplet (AT) loss as:
\begin{equation}
\mathcal{L'}_{AT}  = \frac{1}{N}  \!\!\!\!\!\! \sum_{(I_a, I_p, I_n)} ([\cos(\mathbf{V}_a, \mathbf{V}_n)]_+ - \cos(\mathbf{V}_a,\mathbf{V}_p)
+ 1).
\end{equation}

To remedy the discrepancy between different modalities we have the bi-directional version of AT loss:
\begin{equation}
\begin{aligned}
\mathcal{L}_{AT} &
= \frac{1}{N}  \!\!\! \!\!\!\!\!\! \sum_{(I_a^{rgb}, I_p^{ir}, I_n^{ir})} \!\!\!\!\!\!
\!\!\! ([\cos(\mathbf{V}_a^{rgb}, \mathbf{V}_n^{ir})]_+ - \cos(\mathbf{V}_a^{rgb},\mathbf{V}_p^{ir}) + 1)\\
& + \frac{1}{N}   \!\!\!\!\!\!\!\!\! \sum_{(I_a^{ir}, I_p^{rgb}, I_n^{rgb})} \!\!\!\!\!\!
\!\!\!\!\! ([\cos(\mathbf{V}_a^{ir}, \mathbf{V}_n^{rgb})]_+ - \cos(\mathbf{V}_a^{ir},\mathbf{V}_p^{rgb}) + 1)
.
\end{aligned}
\label{equ_atloss}
\end{equation}

What's more,  as  the derivative of  exponential function $y = e^x$ (which is itself) is larger than 1 and increases exponentially when $x>0$, we make use of this property to accelerate the optimization. Finally, we formulate the Bi-directional Exponential Angular Triplet Loss as:
\begin{equation}
\begin{aligned}
\mathcal{L}_{expAT} &
= \alpha \cdot \frac{1}{N}  \!\!\!\!\!\!  \sum_{(I_a^{rgb}, I_p^{ir}, I_n^{ir})} \!\!\!
\!\!\! e^{([\cos(\mathbf{V}_a^{rgb}, \mathbf{V}_n^{ir})]_+ - \cos(\mathbf{V}_a^{rgb},\mathbf{V}_p^{ir}) + 1)}\\
& +  \beta \cdot \frac{1}{N}  \!\!\!\!\!\!  \sum_{(I_a^{ir}, I_p^{rgb}, I_n^{rgb})} \!\!\!
\!\!\!\!\! e^{([\cos(\mathbf{V}_a^{ir}, \mathbf{V}_n^{rgb})]_+ - \cos(\mathbf{V}_a^{ir},\mathbf{V}_p^{rgb}) + 1)}
,
\end{aligned}
\label{equ_expatloss}
\end{equation}
\noindent here the weights $\alpha$ and $\beta$ are set to be 1 by default.

\noindent \textbf{Discussion of  Euclidean Metric and Cosine Metric.}
Euclidean metric and cosine metric are equivalent for L2-normalized vectors, and the formulation of Euclidean metric $\mathbb{D}$ and cosine metric $\mathbb{C}$ of vectors $\mathbf{X}$ and $\mathbf{Y}$ are connected in math:
\begin{equation}
\mathbb{C}(\mathbf{X},\mathbf{Y}) =1-\cos(\mathbf{X},\mathbf{Y})= 1-\frac{\|\mathbf{X}\|^2+\|\mathbf{Y}\|^2-\mathbb{D}^2(\mathbf{X},\mathbf{Y})}  {2\|\mathbf{X}\| \|\mathbf{Y}\|}.
\end{equation}

However, as the embedding vectors are usually not L2 normalized in Re-ID, Euclidean metric and cosine metric have very different physical significance, and they function differently in the optimization process.
Based on the results of our experiments, the loss functions fail to converge when imposing L2 normalization on the embedding vectors.

\noindent \textbf{Discussion of  the Similarity in Formulation with N-pair Loss.}
N-pair Loss~\cite{npairloss} improves triplet loss by considering more negative pairs in one update.
It adopts the dot product of embedding vectors as distance metric in training and evaluates the algorithm performance with cosine metric in testing.
N-pair loss constructs a mathematical form similar to softmax loss by using \emph{log} and \emph{exp} functions. In our expAT loss, the motivation is different. The expAT Loss aims at facilitating the optimization by using \emph{exp} function taking advantage of the property of derivative of exponential function. A comparable version of N-pair Loss is ``(2+1)-tuplet Loss'' as  suggested in their paper~\cite{npairloss}:

\begin{equation}
\mathcal{L}_{(2+1)-tuplet} = \log(1+e^{\mathbf{V}_a ^\top \mathbf{V}_n -\mathbf{V}_a^\top \mathbf{V}_p}).
\label{equ_npairloss}
\end{equation}

\subsubsection{\textbf{Common Space Batch Normalization}}

\noindent Batch Normalization (BN) is widely adopted in the convolutional layers and enables faster and more stable training of deep neural networks~\cite{BN}.
Generally speaking, BN is a technique that aims to improve the training of neural networks by stabilizing the distributions of layer inputs. This is achieved by introducing a special BN layer to control the first two moments (mean and variance) of feature distributions.

Recent work~\cite{BNnips} points out that BN is able to stabilize the distribution of internal activations and smoothen the optimization landscape.  Inspired by this, we propose to apply a weight-sharing 1-D BN layer on the common feature space to assist AT loss in magnitude stabilizing. At the same time, with common space batch normalization, the channels of embedding vectors are normalized, rescaled and shifted with trainable parameters, which helps recalibrate the channels of embedding vectors.

Here we quickly review the batch normalization layer. For the embedding vector $\mathbf{V}_{raw}$ with $K$ channels obtained from the feature extractor, it is firstly normalized by each channel $\mathbf{V}^k_{raw}$:
\begin{equation}
\begin{aligned}
\hat{\mathbf{V}}^k_{raw} = \frac{\mathbf{V}^k_{raw} - \mathrm E [\mathbf{V}^k_{raw}]}{\sqrt{\mathrm{Var}[\mathbf{V}^k_{raw}]}}, k \in \{1,2,...,K\},
\end{aligned}
\end{equation}
here the statistical expectation $\mathrm E[\cdot]$ and variance $\mathrm{Var} [\cdot]$ are computed in a moving average manner over the training data.

As the normalization operation changes the distribution of features, two trainable parameters, scale and shift terms ($\gamma^k$, $\beta^k$) are introduced for each normalized activation $\hat{\mathbf{V}}^k_{raw}$ to make sure that the batch normalization can degrade to the identity transformation (when $\gamma^k = \sqrt{\mathrm{Var}[\mathbf{V}^k_{raw}]}$, $\beta^k = \mathrm E[\mathbf{V}^k_{raw}]$).

\textit{Notably},
the shift term $\beta^k$ is removed from CSBN to avoid center drift of feature space.
Thus with the common space batch normalization operator ($\mathrm{CSBN}$), each channel  $\mathbf{V}^k$ of the final embedding vector is calculated as:
\begin{equation}
\begin{aligned}
\mathbf{V}^k &= \mathrm{CSBN}(\mathbf{V}_{raw}^k) \\
&= \gamma^k \hat{\mathbf{V}}^k_{raw}, k \in \{1,2,...,K\}.
\end{aligned}
\label{equ_csbn}
\end{equation}

The following experiment section indicates that the expAT loss needs to work with the CSBN to achieve large performance gain.

\subsubsection{\textbf{Identity Loss}}
The identity loss $\mathcal{L}_{id}$ is a softmax function based cross entropy loss   widely used in classification tasks. It indicates the distance between what the model believes the output distribution should be, and what the original distribution is.

A fully-connected layer $\mathbb{F}$ is deployed to map the embedding vector $\mathbf{V}$ from common feature space to the probability distribution space of different classes~(person identities), followed by the identity loss.

Formally, given an image $I$ and its embedding vector $\mathbf{V}$, we denote $y$ as ground truth label, $p$ as predicted probability vector and $p_i$ indicates the predicted probability of class $i$~($C$ classes in total), identity loss is:
\begin{equation}
\mathcal{L}_{id} = -  \sum_{i=1}^{C} q_i log(p_i),
q_i=\left\{
\begin{aligned}
1, i = y \\
0, i \neq y \\
\end{aligned}
\right.
,
\end{equation}
\begin{equation}
p_i =   \frac{e ^ {\mathbf{V}^{fc}_i}}{\sum_{k=1}^{C} {e^{\mathbf{V}^{fc}_k}}},
\end{equation}

\noindent here $\mathbf{V}^{fc}_i$ is the corresponding element of class $i$ in the column vector $\mathbf{V}^{fc}$ computed by the fully-connected layer:
\begin{equation}
\mathbf{V}^{fc} = \mathbb{F}(\mathbf{V}).
\label{equ_linearlayer}
\end{equation}

Considering both modalities,  after extracting the embedding vectors ($\mathbf{V}^{rgb}$, $\mathbf{V}^{ir}$) of an RGB image $I^{rgb}$ and an IR image $I^{ir}$ of the same person, we denote $y$ as the ground truth label, $p_i^{rgb}$ and $p_i^{ir}$ indicate the probabilities of class $i$ predicted from RGB image and IR image separately, the identity loss $\mathcal{L}_{id}$ is formally written as:
\begin{equation}
\mathcal{L}_{id} = - \sum_{i=1}^{C} (q_i (log(p_i^{rgb}) + log(p_i^{ir}))), \\
q_i=\left\{
\begin{aligned}
1, i = y \\
0, i \neq y \\
\end{aligned}
\right.
,
\label{equ_cid}
\end{equation}
\begin{equation}
\begin{aligned}
p_i^{rgb} &= \frac{e ^ {(\mathbb{F}(\mathbf{V}^{rgb}))_i}}{\sum_{k=1}^{C} {e^{(\mathbb{F}(\mathbf{V}^{rgb}))_k}}},\\
p_i^{ir} &= \frac{e ^ {(\mathbb{F}(\mathbf{V}^{ir}))_i}}{\sum_{k=1}^{C} {e^{(\mathbb{F}(\mathbf{V}^{ir}))_k}}}.
\end{aligned}
\end{equation}

By directly summing the two constraints mentioned above, we come up with the final hybrid loss function in the proposed framework:
\begin{equation}
\mathcal{L} = \mathcal{L}_{expAT} + \mathcal{L}_{id}.
\end{equation}

As for optimization method, the  first-order gradient-based optimization algorithm is employed to optimize the parameters of model.   The overall algorithm of training the proposed model is presented in Algorithm~\ref{alg}.

\renewcommand{\algorithmicrequire}{ \textbf{Input:}} 
\renewcommand{\algorithmicensure}{ \textbf{Output:}} 
\begin{algorithm}
	\caption{Training of the Proposed Network}
	\label{alg}
	\begin{algorithmic}
		\REQUIRE ~~\\
		Training set, initialized network parameters, learning rate		
		\REPEAT
		\STATE Get training samples: \{$I_a^{rgb}$,  $I_a^{ir}$, $I_p^{rgb}$, $I_p^{ir}$, $I_n^{rgb}$, $I_n^{ir}$, $y$\}, $y$ is identity label of anchor. \\
		\STATE Calculate the embedding vectors after common space batch normalization with Equation~\ref{equ_extractor}: \{$\mathbf{V}_a^{rgb}$, $\mathbf{V}_a^{ir}$, $\mathbf{V}_p^{rgb}$, $\mathbf{V}_p^{ir}$, $\mathbf{V}_n^{rgb}$,$\mathbf{V}_n^{ir}$\}.
		\STATE Calculate the bi-directional expAT loss with Equation~\ref{equ_expatloss}: $\mathcal{L}_{expAT}(\mathbf{V}_a^{rgb}, \mathbf{V}_a^{ir}$, $\mathbf{V}_p^{rgb}$, $\mathbf{V}_p^{ir}$, $\mathbf{V}_n^{rgb}$,$\mathbf{V}_n^{ir})$.
		\STATE Calculate the identity loss with Equation~\ref{equ_cid}: $\mathcal{L}_{id}( \mathbf{V}_a^{rgb}, \mathbf{V}_a^{ir}, y)$.
		\STATE Calculate the total loss: $\mathcal{L} = \mathcal{L}_{expAT} + \mathcal{L}_{id}$.
		\STATE Backward and update network parameters.
		\UNTIL{\textbf{convergence}}
		\ENSURE ~~\\
		Trained network parameters
	\end{algorithmic}
\end{algorithm}

\begin{table*}[h]
	\centering
	\caption{Ablation study on the Single-shot setting of SYSU-MM01 dataset. ``CSBN'' means using common space batch normalization. ``EL'' means adding an additional embedding layer. ``exp'' means using the expAT Loss. Rank-1,10 and 20 accuracy~(\%) as well as  mAP (\%) are listed. The best results are highlighted in bold.}
	\scalebox{0.9}{
		\setlength{\tabcolsep}{3mm}
		{
			\begin{tabular}{cccccc|cccc}
				\bottomrule
				\multirow{2}{*}{ID Loss}& 	
				\multirow{2}{*}{Triplet Loss}& 	\multirow{2}{*}{AT Loss} &	\multirow{2}{*}{EL} & 	\multirow{2}{*}{CSBN} & 	\multirow{2}{*}{exp}  &\multicolumn{4}{|c}{Single-shot}
				\\
				\cline{7-10}
				&&&&&& Rank-1 & Rank-10 & Rank-20 & \multicolumn{1}{c}{mAP}
				\\
				\hline
				\checkmark&   &  & & && 7.40& 39.16&56.34&11.46
				\\
				\checkmark&   &  & &\checkmark & & 19.92 & 39.69 & 61.40 & 17.80
                \\
				\checkmark&\checkmark &  &  & &&26.15& 65.05& 76.56&25.57
				\\
				\checkmark&\checkmark  & &   &\checkmark&& 25.91& 64.78& 77.31& 25.88
				\\
				\checkmark&& \checkmark &  & &&  12.95& 54.74&  70.97& 17.09
				\\
				\checkmark && \checkmark&  \checkmark &\checkmark&&31.12&70.40 & 82.87&33.01
				\\
				\checkmark&& \checkmark &  &\checkmark&& 37.00&75.82&85.68&37.28
				\\
				\checkmark&& \checkmark &  &\checkmark&\checkmark & \textbf{38.57} &  \textbf{76.64} &  \textbf{86.39} & \textbf{38.61}
				\\
				\toprule
	\end{tabular}}}
	\label{tab_q1}
\end{table*}

\section{Experiments}
In this part, to validate our method, we perform introspective study and measure against latest baselines on RGB-IR Re-ID. Experiments are designed to answer the following two questions:

\noindent \textbf{Q1 Ablation Study:} How do the components contribute to the performance?

\noindent \textbf{Q2 Peer Comparison:} How is the proposed framework compared with the state-of-the-art methods?

\subsection{Benchmark}
The experiments are conducted with three prevalent datasets for RGB-IR Re-ID (SYSU-MM01~\cite{SYSUMM01}), RGB-thermal Re-ID (RegDB~\cite{nguyen2017}) and RGB Re-ID (Market-1501~\cite{market} and DukeMTMC-reID~\cite{zheng2017unlabeled}). For performance evaluation, rank-$k$ accuracy and mean average precision~(mAP) are used.
\subsubsection{SYSU-MM01 Dataset}
The SYSU-MM01 is the first public dataset for RGB-IR Re-ID. It is a large-scale dataset including RGB and IR images of 491 identities from 6 cameras, providing in total 287,628 RGB images and 15,792 IR images.

\noindent\textbf{Evaluation Protocol.}
According to the official evaluation protocol, in the all-search mode of SYSU-MM01 dataset, RGB cameras 1, 2, 4 and 5 are for gallery set while IR cameras 3 and 6 are for query set. For every identity under an RGB camera, the dataset randomly chooses one/ten image(s) of the identity to form the gallery set for single-shot/multi-shot setting. As for the query set, all IR images are used. Given a query image, matching is conducted by computing similarities between the query image and gallery images.
Notice that matching is conducted between cameras in different locations, thus the query images of camera 3 skip the gallery images of camera 2 since they are at the same location. In the training stage, there is a fixed split with 296 identities for training and 99 for validation.
In the testing stage, under all-search single-shot/multi-shot setting, there are 96 persons, with 3,803 IR images as query and 301/3010 randomly selected RGB images as gallery set.
Evaluation is repeated 10 times with different random splits of gallery set to compute the statistically stable performance indicators~\cite{SYSUMM01}.

\subsubsection{RegDB Dataset}
RegDB dataset~\cite{nguyen2017} is a public dataset for RGB-thermal person re-identification. It contains 8,240 images of 412 people captured by one RGB camera and one thermal camera. Each identity has 10 RGB images (visible light) and 10 thermal images (infrared light).

\noindent\textbf{Evaluation Protocol.} We follow an evaluation practice adopted by many researchers~\cite{tone}: The identities are randomly split (according to code provided by the authors of \cite{tone}) into two halves, one for training and the other for testing. In the
testing stage, the images from one modality were used as the gallery set while the ones from the other modality as the probe set.
There are two query settings: ``Thermal to RGB'' and ``RGB to Thermal''. In ``Thermal to RGB'', thermal images form query set while RGB images form gallery set, while in ``RGB to Thermal'' the query set and gallery set interchange with each other.

\subsubsection{Market-1501 Dataset}
The Market-1501  dataset is a popular single-modality RGB Re-ID dataset. It is collected at some daytime  outdoor  scenes in Tsinghua University. A total of six cameras are used, including 5 high-resolution cameras (1280$\times$1080) and one low-resolution (720$\times$576) camera. Field-of-view overlap exists among different cameras. Overall, this dataset contains 32,668 annotated bounding boxes of 1,501 identities. Each person is captured by at least two cameras to make the cross-view retrieval possible. The bounding boxes are provided by a detection algorithm DPM~\cite{DPM}, which brings lots of misalignment and part missing. Thus it requires the Re-ID algorithm to overcome more intensive intra-class variation and inter-class similarity in the detected images.

\noindent\textbf{Evaluation Protocol.} The Market-1501 splits images of 750 identities as training set and those of 751 identities as testing set. At testing phase, only one image is selected as query for each identity from every camera (single-shot setting). Totally there are 3,368 query images.

\subsubsection{DukeMTMC-reID Dataset}
The DukeMTMC-reID dataset  is collected from DukeMTMC~\cite{duke} and contains 36,411 bounding boxes of 1,812 identities shot by 8 cameras.

\noindent\textbf{Evaluation Protocol.} The dataset consists of 16,522 training images of 702 identities and 19,889 testing images of 702 identities. There are also 408 distractor IDs which only appear in one camera. In the testing set, there are  2,228 query images and 17,661 gallery images.

\begin{figure}[t]
	\centering
	\includegraphics[scale=0.33]{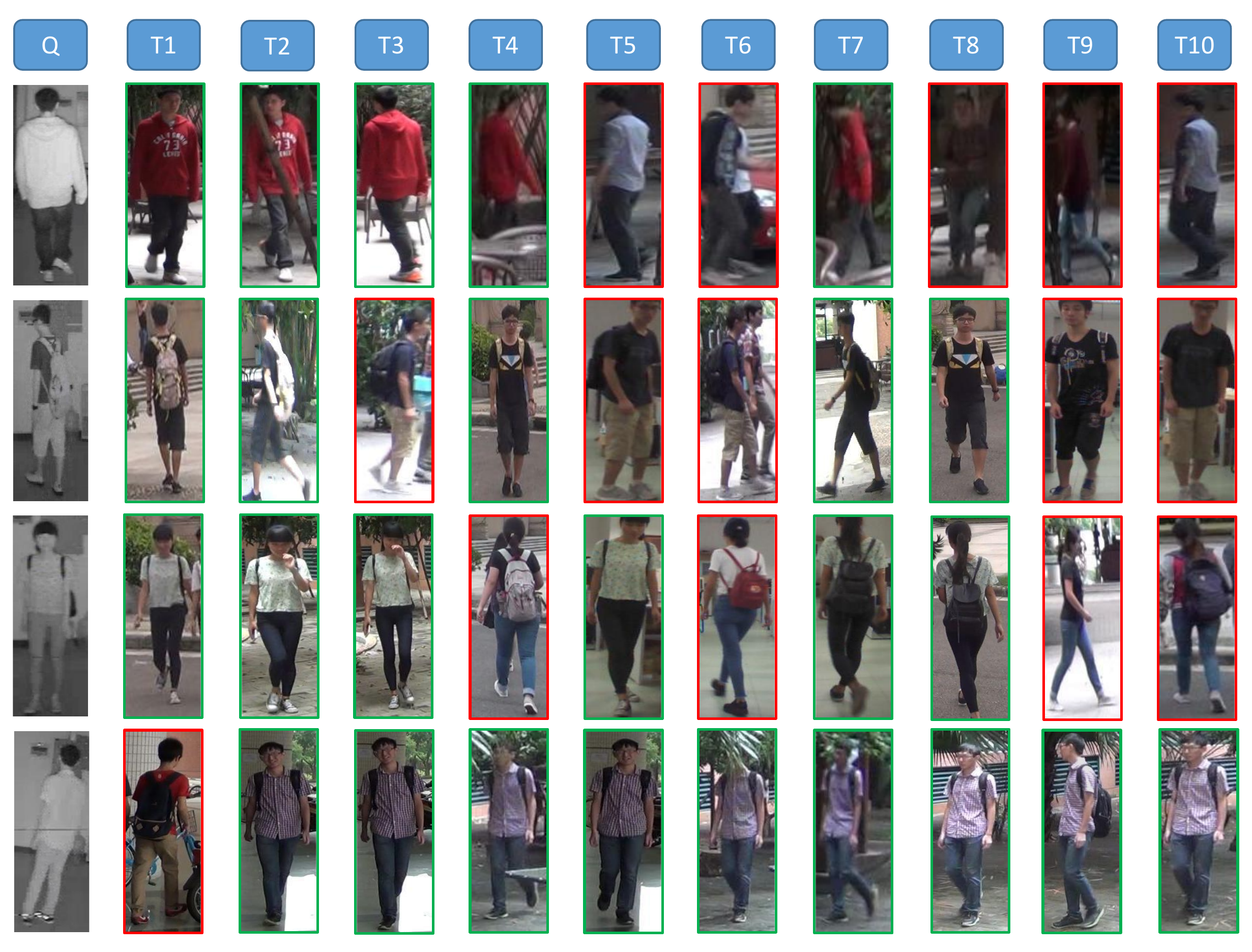}
	\caption{Retrieval results of expAT Loss on SYSU-MM01 dataset under all-search multi-shot setting.
    In each row, top-\emph{k} images are arranged in ascending order according to their distances with the query on the left.
    The true and false matches are in the green and red boxes, respectively.}
	\label{img_res_vis_sysu}
\vspace{-1.em}
\end{figure}

\subsection{Implementation details}
For the experiments of baseline model and the proposed methods on the SYSU-MM01 dataset, the algorithms are implemented with Pytorch 1.0. Mini-batch size is set as 8. Each image is loaded as raw pixel values in range $[0, 1]$ with three channels~(for IR images, the values of three channels are identical), and then resized to $384 \times 128$ as well as normalized by subtracting 0.485, 0.456, 0.406 and dividing by 0.229, 0.224, 0.225 for three channels respectively.
ADAM optimizer~\cite{adam} with Warmup~\cite{warmup} strategy is utilized for optimization, with the  initial learning rate set as $3\times10^{-4}$ and decayed by 0.1 at 10,000 and 20,000 step. Totally there are 30,000 iteration steps in training.  Margin $\xi$ in the triplet loss is set as 0.3 for anchors of both modalities.  To ease the overfitting problem in training, the Label Smooth~\cite{labelsmooth}  and the RandomErasing~\cite{randomerase} strategies are adopted in training. The label smoothing parameter is set as 0.1 and the random erasing probability is set as 0.5. Kaiming initialization~\cite{kaiminginit} is applied to the fully-connected layers. In testing, feature distance is calculated by Euclidean metric.

\noindent \textbf{Batch Sampling Strategy.} We adopt a special sampling strategy to fit the implementation of cross-modality constraints. First, we select an RGB image and an IR image of the same person as image pair. In each mini-batch with batch size $N$, $N$ anchor image pairs are randomly selected from the whole training set. Then, for each anchor RGB image, we randomly choose a positive IR image and a negative IR image from the training set excluding the anchor IR image; for each anchor IR image, we randomly choose a positive RGB image and a negative RGB image from the training set excluding the anchor RGB image. Thus we obtain $N$ tuples with totally $N\times 2\times 3$ images in each mini-batch. In each tuple, the anchor images of the same person from different domains are used for computing identity loss, while all images are used for calculating the ranking loss. All training images will be traversed with the randomly sampling strategy in training.

\begin{table}
	\centering
	\caption{Comparison among the variants of the proposed method (expAT loss).}
	\scalebox{1}{
		\setlength{\tabcolsep}{2.5mm}
		{
			\begin{tabular}{c|cccc}
				\bottomrule
				& Rank-1 & Rank-10 & Rank-20 & \multicolumn{1}{c}{mAP}
				\\
				\hline
				CSBN\_Full & 35.19& 75.38 & 86.00 & 36.00
				\\
				CSBN\_None & 32.88 & 72.44 & 84.76 & 34.45
				\\
				uni-directional & 21.62 & 49.29 & 60.71 & 24.88
				\\
				\hline
				Proposed & 38.57 &  76.64 & 86.39 & 38.61
				\\
				\toprule
	\end{tabular}}}
	\label{tab_variant}
	\vspace{-1.2em}
\end{table}

\begin{table}
	\centering
	\caption{Comparison between the naive bi-directional cosine distance based triplet loss and the AT and expAT loss. Rank-1,10 and 20 accuracy~(\%) as well as mAP (\%) are listed. }
	\scalebox{1}{
		\setlength{\tabcolsep}{2.5mm}
		{
			\begin{tabular}{c|cccc}
				\bottomrule
				& Rank-1 & Rank-10 & Rank-20 & \multicolumn{1}{c}{mAP}
				\\
				\hline
				Naive Cosine Triplet &36.51& 74.23& 83.46&36.33
				\\
				AT (Ours)&37.00&75.82&85.68&37.28
				\\
				expAT (Ours) & 38.57 &  76.64 & 86.39 & 38.61
				\\
				\toprule
	\end{tabular}}}
	\label{tab_q13}
	\vspace{-1.2em}
\end{table}

\begin{table}
	\centering
	\caption{Comparison of different margins $\xi$ in expAT loss. Rank-1,10 and 20 accuracy~(\%) as well as mAP (\%) are listed. }
	\scalebox{1}{
		\setlength{\tabcolsep}{2.5mm}
		{
			\begin{tabular}{c|cccc}
				\bottomrule
			 $\xi$	 & Rank-1 & Rank-10 & Rank-20 & \multicolumn{1}{c}{mAP}
				\\
				\hline
				0.0 & 18.23 & 46.57 & 57.94 & 15.96
				\\
				0.5 & 39.35 & 75.66 & 85.27 & 38.23
				\\
				1.0 & 38.57 &  76.64 & 86.39 & 38.61
				\\
                1.5 & 38.26 & 75.14 & 84.49 & 37.07
                \\
                2.0 & 33.74 & 72.58 & 84.68 & 35.09
                \\
				\toprule
	\end{tabular}}}
	\label{tab_margin}
	\vspace{-1.2em}
\end{table}

\begin{table}[h]
	\centering
	\caption{Ablation study of different training practices. ``WU'' means using Warmup~\cite{warmup}. ``LS'' means using  Label Smooth~\cite{labelsmooth}. ``RE'' means using the RandomErasing~\cite{randomerase}. Rank-1,10 and 20 accuracy~(\%) as well as  mAP (\%) are listed. }
	\scalebox{0.9}{
		\setlength{\tabcolsep}{3mm}
		{
			\begin{tabular}{ccc|cccc}
				\bottomrule
				\multirow{2}{*}{LS}& 	
				\multirow{2}{*}{RE}& 	
                \multirow{2}{*}{WU}
                &\multicolumn{4}{|c}{Single-shot}
				\\
				\cline{4-7}
				&&& Rank-1 & Rank-10 & Rank-20 & \multicolumn{1}{c}{mAP}
				\\
				\hline
				-&  - &-  & 30.70 & 71.66 & 83.27 & 32.89
				\\
				\checkmark& -  & -  & 33.69 & 70.77 & 82.67& 34.86
                \\
                -& \checkmark & -  & 33.02  & 74.37  & 84.67  & 35.07
                \\
                -& - & \checkmark  & 32.40 & 69.50 & 80.73 & 34.48
                \\
				\checkmark&\checkmark  &-& 33.93 & 71.13 & 83.17 & 34.97
				\\
				\checkmark  &-& \checkmark & 31.97 &68.15 &79.77 &33.18
				\\
				-& \checkmark & \checkmark & 34.48 & 75.46 & 85.81 & 36.16
				\\
				\checkmark&\checkmark  &\checkmark & 38.57 & 76.64 & 86.39 & 38.61
				\\
				\toprule
	\end{tabular}}}
	\label{tab_q14}
	\vspace{-1.3em}
\end{table}

\subsection{Ablation Study}
In this subsection, we study the influence of each component in the proposed framework by comparing the quantitative results of its variants and analyzing the visualization of feature space.  All studied models share the same feature extractor structure and training skills. The experiments are conducted on the SYSU-MM01 dataset with all search single-shot setting if not otherwise stated.

The primary experiment results are listed in the Table~\ref{tab_q1}. ``CSBN'' means using common space batch normalization. ``EL'' means adding an additional embedding layer. ``exp'' means using the expAT loss.

\begin{table*}[h]
	\centering
	\caption{Peer comparison on the SYSU-MM01 dataset.  Rank-1 accuracy~(\%) and  mAP (\%) are listed.  ``Metric'' is the metric used in testing phase. "-" means not provided or not applicable. The best results are highlighted in bold  and the second best are underlined.}
	\scalebox{0.8}{
		\setlength{\tabcolsep}{4mm}
		{
			\begin{tabular}{c|c|cccc|cccc}
				\bottomrule
				\multirow{2}{*}{Method} & \multirow{2}{*}{Metric} &\multicolumn{4}{c|}{Single-shot} & \multicolumn{4}{c}{Multi-shot}
				\\
				\cline{3-10}
				&& Rank-1 & Rank-10 & Rank-20  & mAP & Rank-1 & Rank-10 & Rank-20  & mAP
				\\
				\hline
				\multirow{8}{*}{HOG~\cite{hog}} & Euclidean & 2.76 &18.25&31.91 & 4.24 & 3.82 & 22.77 & 37.63 &2.16
				\\
				& KISSME~\cite{kissme} & 2.12 & 16.21& 29.13& 3.53 &2.79 &18.23 &31.25& 1.96
				\\
				& LFDA~\cite{lfda} &2.33 & 18.58& 33.38& 4.35& 3.82& 20.48 &35.84 &2.20
				\\
				& CCA~\cite{cca} & 2.74 &18.91& 32.51& 4.28 &3.25& 21.82& 36.51& 2.04
				\\
				& CDFE~\cite{cdfe} & 2.09 &16.68& 30.51& 3.75& 2.47& 19.11& 34.11& 1.86
				\\	
				& GMA~\cite{gma} &1.07 &10.42& 20.91& 2.52 &1.03& 10.29& 20.73 &1.39
				\\				
				& SCM~\cite{scm}  &1.86 &15.16 &28.27& 3.57 &2.40& 17.45& 31.22 &1.66
				\\	
				& CRAFT~\cite{hiphop} &2.59 &17.93 &31.50& 4.24 &3.58 & 22.90& 38.59 &2.06
				\\	
				\hline
				\multirow{8}{*}{LOMO~\cite{lomo}}& Euclidean & 1.75 & 14.14 &26.63& 3.48& 1.96& 15.06& 27.30& 1.85
				\\
				&KISSME~\cite{kissme}& 2.23& 18.95 &32.67& 4.05& 2.65 &20.36& 34.78& 2.45
				\\
				&LFDA~\cite{lfda} &2.98& 21.11 &35.36 &4.81 &3.86 &24.01 &40.54 &2.61
				\\
				&CCA~\cite{cca} &2.42 &18.22 &32.45 &4.19 &2.63 &19.68 &34.82 &2.15
				\\
				&CDFE~\cite{cdfe} &3.64 &23.18 &37.28 &4.53 &4.70 &28.23 &43.05 &2.28
				\\
				&GMA~\cite{gma} &1.04 &10.45 &20.81 &2.54 &0.99 &10.50 &21.06 &1.47
				\\
				&SCM~\cite{scm} &1.54 &14.12 &26.27 &3.34 &1.66 &15.17 &28.41 &1.57
				\\
				&CRAFT~\cite{hiphop} &2.34 &18.70 &32.93 &4.22 &3.03 &21.70 &37.05 &2.13
				\\
				\hline
				Deep Zero-Padding~\cite{SYSUMM01}& Euclidean & 14.80 &54.12& 71.33 &15.95 &19.13 &61.40 &78.41 &10.89
				\\
				HCML~\cite{tone}& Euclidean & 14.32 &53.16 &69.17&  16.16 & - & -&-& -
				\\
				BDTR~\cite{VT}& Euclidean & 17.01 &55.43& 71.96& 19.66 & - & - & -& -
				\\
				cmGAN~\cite{cmgan}& Euclidean & 26.97 & 67.51 & 80.56& 27.80  & 31.49&  \underline{72.74}& 85.01  & 22.27
				\\
				D$^2$RL~\cite{GANCVPR19}& Euclidean & 28.90 & 70.60 & 82.40 & 29.20 & - & - & - & -
				\\
				DGD~\cite{DGD} & Euclidean & 22.77 & 65.90 & 80.66 & 23.76 & 27.81 & 72.46 &  \underline{85.95} & 17.30
				\\
				MSR~\cite{msr2019tip} & Euclidean &  \underline{37.35} & \textbf{83.40} & \textbf{93.34} &  \underline{38.11} &  \underline{43.86} & \textbf{86.94} & \textbf{95.68} &  \underline{30.48}
				\\
				\hline
				BNNeck~\cite{luohao}& Euclidean & 30.11 & 71.25 & 83.25 & 32.03 & 34.60 & 64.49 & 75.15 & 25.26
				\\				
				Cosine softmax loss~\cite{wojke2018} & Euclidean  & 12.01 & 37.58 & 47.03 & 12.28 & 15.19 & 41.18 & 51.85 & 7.57
				\\		
                N-pair ((2+1)-tuplet) Loss~\cite{npairloss}& Cosine & 26.69 & 64.96 & 77.07 & 26.90 & 33.18 & 62.46 & 73.01 & 20.65
                \\
				AT Loss (\textbf{Ours})& Euclidean &  37.00 & 75.82&  85.68&  37.28 & 43.34 &  71.23&  79.49 &  30.42
				\\
				expAT Loss (\textbf{Ours})& Euclidean  & \textbf{38.57} &  \underline{76.64} &  \underline{86.39} & \textbf{38.61} & \textbf{44.71} & 69.82 & 77.87 & \textbf{32.20}
				\\
				\toprule
	\end{tabular}}}
	\label{tab_q2}
\end{table*}

\subsubsection{Study of CSBN}
As shown in Table~\ref{tab_q1}, we evaluate the influence of common space batch normalization~(CSBN) on ID loss, triplet loss (bi-directional) and AT loss. First,  the results of baseline model with Identity loss only~(ID Loss) are shown.  Then, we combine the ID loss with bi-directional triplet loss~(ID+Triplet). ID+Triplet achieves competitive performance, but it is not compatible with CSBN, leading to a slight performance decrease of rank-1 accuracy (-0.24\%). It is because normalization operation reshapes the distribution of embedding vectors and impairs the Euclidean metric based constraints among samples.
On the contrary, AT and ID loss both benefit from CSBN. AT loss achieves a remarkable performance gain (rank-1: +24.05\%, mAP: +20.19\% in single-shot setting) compared with not using CSBN. The reason is that CSBN stabilizes the magnitudes of embedding vectors, and helps recalibrate the directions.

To further discuss the function of CSBN, we study the variants of it and show their performance in Table~\ref{tab_variant}. The original CSBN is formulated as Equation~\ref{equ_csbn}. First we add the shift parameter $\beta^k$ on CSBN (CSBN\_Full). This setting is identical with the popular batch normalization used between convolutional layers.  And then we remove all the trainable parameters $\gamma^k$, $\beta^k$ (CSBN\_None). In this way, the function of CSBN is only stabilizing the magnitudes of embedding vectors in common feature space. Quantitative results show that both CSBN\_Full and CSBN\_None are not as good as CSBN.

As L2 normalization (L2-norm) is a stringent constrain of the magnitudes of feature embedding vectors, we also try to replace CSBN with L2-norm on the experiments of AT Loss and expAT Loss.
However, experiments find that with L2-norm the loss functions could not converge.

\subsubsection{Study of Loss Function}
We show the performance of naive cosine triplet loss $\mathcal{L}_{cos}$ (with bi-directional manner and CSBN)  and compare with our methods in Table \ref{tab_q13}, which manifests the effectiveness of our improvement of  loss function.

To study how the margin $\xi$ in expAT loss (set to be 1 in Equation~\ref{equ_expatloss}) influences the performance, we try different margins and list the results in Table~\ref{tab_margin}.
The experiment results show  that, when $\xi=1$, the algorithm shows the best performance considering mAP.
We argue that when $\xi<1$, the exponent of Equation~\ref{equ_expatloss} may be negative,  which shrinks the gradients when training; when  $\xi>1$, the loss may be unnecessarily large and influence the effectiveness of identity loss;
$\xi=1$ is just enough to keep the exponent non-negative.
To make the hyper-parameters simple, we set $\xi=1$ in expAT loss.

\begin{table}
	\centering
	\caption{Comparison of different ratios of $\alpha$ and $\beta$ ($\alpha:\beta$) in expAT Loss. Rank-1,10 and 20 accuracy~(\%) as well as mAP (\%) are listed. }
	\scalebox{1}{
		\setlength{\tabcolsep}{2.5mm}
		{
			\begin{tabular}{c|cccc}
				\bottomrule
			 $\alpha:\beta$	 & Rank-1 & Rank-10 & Rank-20 & \multicolumn{1}{c}{mAP}
				\\
				\hline
				1:1 & 38.57 &  76.64  & 86.39  & 38.61
				\\
				1:2 & 36.59 &  75.85 &  86.07 &  38.16
				\\
				2:1 & 38.23 & 76.67 & 86.94 & 38.64
				\\
                1:3 & 36.52 & 75.29 & 86.14 & 37.54
                \\
                3:1 & 36.50 & 76.95 & 86.64 & 38.23
                \\
				\toprule
	\end{tabular}}}
	\label{tab_lossweight}
	\vspace{-1.2em}
\end{table}

As the IR images and RGB images contain different appearance information, different ratios of the RGB term and IR term in expAT Loss should influence the performance.
In this experiment, we fix the sum of the $\alpha$ (weight of IR term) and $\beta$ (weight of RGB term) in expAT Loss to 2 and try different ratios. The results are listed in Table~\ref{tab_lossweight}. When $\alpha:\beta = 2:1$, the algorithm yields slightly better performance regarding mAP.
However, when the weights $\alpha$ and $\beta$   get more unbalanced, the performance becomes worse.

\subsubsection{Study of Backbone Model}
As discussed in the Section~\ref{sec_network}, the effect of an additional embedding layer (EL) is evaluated based on the model with ID loss and AT loss. Experiment results in Table~\ref{tab_q1} demonstrate a performance decrease (-5.88\% for rank-1 accuracy and -4.27\% for mAP). Thus an additional embedding layer is unsuitable for this task.

Moreover, we investigate how different training practices in the optimization process, including Warmup~\cite{warmup}, Label Smooth~\cite{labelsmooth} and  RandomErasing~\cite{randomerase}, influence the performance of expAT.
The results are listed in Table~\ref{tab_q14}. Without using any of these practices, the Rank-1 accuracy and mAP of bi-directional expAT Loss with CSBN are 30.70\% and 32.89\%.
Notably, if we only use two or one of them, the performance will be much worse than using all of them.

\subsubsection{Study of bi-directional manner}
As discussed in the Section~\ref{sec_constraints}, we adopt a bi-directional manner~\cite{VT} as default in the triplet-like loss functions. When the anchor term is sampled from RGB domain, the positive and negative terms are sampled from IR domain, and vice versa. To evaluate the influence of this manner, we compare the experiment results of using uni-directional manner and using bi-directional manner on expAT loss in Table~\ref{tab_variant}.
When using the uni-directional manner, the anchor, positive and negative terms  in a triplet are all from the same modality.
As shown in Table~\ref{tab_variant}, bi-directional manner brings a performance improvement of 16.95\% and 13.73\% correspondingly for rank-1 accuracy and mAP.

\subsection{Overall Effect and Visualization Analysis}
Compared with baseline model, the proposed single-stream framework with expAT loss rises the rank-1 accuracy from 7.40\% to 38.57\% and the mAP from 11.46\% to 38.61\% with exactly the same backbone structure and hyper-parameter setting.
Moreover, 2-D visualization of embedding vectors generated from the testing set of SYSU-MM01 dataset  is shown in the Figure~\ref{img_pca}.
It is  observed that the embedding vectors with expAT loss are apparently more separable in common feature space than those with Triplet loss.
Both the quantitative and qualitative results validate the effectiveness of our method.

We provide some cross-modality retrieval results on SYSU-MM01 dataset under all-search multi-shot setting in Figure~\ref{img_res_vis_sysu}.
As infrared images contain little color information, the model gets confused when dealing with images containing similar contours but different colors (e.g. T5 and T6 in the first row, T3 and T5 in the second row of the figure~\ref{img_res_vis_sysu}).
The texture information in infrared modality is also inferior because of the imaging characteristic of surveillance infrared cameras.
As shown in the query image of fourth row, the plaid of shirt is not recognizable, which leads to the matching failure at the corresponding top-1 result.

\subsection{Peer Comparison}

\subsubsection{Comparison on RGB-IR Re-ID}
In this subsection, we compare our method with other RGB-IR Re-ID algorithms on SYSU-MM01 dataset under all search setting.

For the traditional hand-crafted feature based methods, we show results of HOG~\cite{hog} and LOMO~\cite{lomo}~(collected from \cite{SYSUMM01})
with different metric learning methods (KISSME~\cite{kissme}, LFDA~\cite{lfda}, CCA~\cite{cca}, CDFE~\cite{cdfe}, GMA~\cite{gma}, SCM~\cite{scm}, CRAFT~\cite{hiphop}).

With the development of neural networks, deep learning based methods show even stronger performance in this field.
For the two-stream models, we have
HCML~\cite{tone}, BDTR~\cite{VT} DGD~\cite{DGD} and MSR~\cite{msr2019tip}. For the single-stream models, we compare with deep zero-padding~\cite{SYSUMM01}, D$^2$RL~\cite{GANCVPR19} as well as cmGAN~\cite{cmgan}.

More related to our work, cosine softmax loss~\cite{wojke2018} has been introduced to single-modality Re-ID task, we re-implement it  based on our  backbone and training skills.
N-pair loss~\cite{npairloss} utilizes the dot product of embedding vectors as distance metric in training and evaluates the algorithm performance with cosine metric in testing.
We implement the bi-directional simplified version of it, which considers only one positive pair and one negative pair in an iteration with bi-directional manner for fair comparison ((2+1)-tuplet loss). It takes more iteration steps for n-pair loss to converge so the total iteration number is set to be 80,000, the learning rate is set to be decayed by 0.1 at 20,000 and 40,000 step.
We also implement a bi-directional version of another improved hybrid loss (ID Loss + Triplet Loss) proposed for RGB Re-ID~(BNNeck~\cite{luohao}) with the same backbone and training skills.
Different from our common space batch normalization, BNNeck  only puts batch normalization on the softmax loss but not on the triplet loss.
BNNeck~\cite{luohao} suggests a bag of tricks (BoT): Random Erasing~\cite{randomerase}, Warm Up~\cite{warmup} and Label Smooth~\cite{labelsmooth}.
To study how the BoT influence the compared algorithms, we carry out ablation study and show the results in Table~\ref{tab_peer_bot}.
The proposed method yields stronger results with or without BoT.

\begin{table}[h]
	\centering
	\caption{Ablation study of ``BoT'' on SYSU-MM01 dataset. ``BoT'' means using Warm Up,  Label Smooth and Random Erasing. Rank-1,10 and 20 accuracy~(\%) as well as  mAP (\%) are listed. }
	\scalebox{0.7}{
		\setlength{\tabcolsep}{3mm}
		{
			\begin{tabular}{c|c|cccc}
				\bottomrule
				\multirow{2}{*}{Algorithm}& 	
                \multirow{2}{*}{BoT}
                &\multicolumn{4}{c}{Single-shot}
				\\
				\cline{3-6}
				&& Rank-1 & Rank-10 & Rank-20 & \multicolumn{1}{c}{mAP}
				\\
				\hline
                Cosine Softmax Loss~\cite{wojke2018} & -  & 4.29  & 20.43  & 29.25  & 5.98
                \\
				N-pair ((2+1)-tuplet) Loss~\cite{npairloss} & - &  21.89  &  64.10 &77.84&24.62
				\\
				BNNeck~\cite{luohao}  & - & 28.66 & 69.24 & 81.08 & 30.33
                \\
                expAT  (\textbf{ours}) & -  & 30.70  & 71.66 & 83.27  & 32.89
                \\
				\hline
                Cosine Softmax Loss~\cite{wojke2018} & \checkmark  & 12.01 & 37.58 & 47.03 & 12.28
                \\
				N-pair ((2+1)-tuplet) Loss~\cite{npairloss} & \checkmark & 26.69 & 64.96 & 77.07 & 26.90
				\\
				BNNeck~\cite{luohao} & \checkmark & 30.11 & 71.25 & 83.25 & 32.03
				\\
                expAT (\textbf{ours}) & \checkmark  & 38.57 & 76.64 & 86.39 & 38.61
                \\
				\toprule
	\end{tabular}}}
	\label{tab_peer_bot}
	\vspace{-1.em}
\end{table}

\begin{table}[h]
	\centering
	\caption{Peer comparison on the RegDB dataset with ``Thermal to RGB'' and ``RGB to Thermal'' query settings.  Rank-1 accuracy~(\%) and  mAP (\%) are listed. The best results are highlighted in bold  and the second best are underlined.}
	\scalebox{0.9}{
		\setlength{\tabcolsep}{3mm}
		{
			\begin{tabular}{c|cc|cc}
				\bottomrule
				\multirow{2}{*}{Method}&\multicolumn{2}{c|}{Thermal to RGB} &\multicolumn{2}{c}{RGB to Thermal}
				\\
				\cline{2-5}
				& Rank-1   & mAP & Rank-1   & mAP
				\\
				\hline
				Deep Zero-Padding~\cite{SYSUMM01} & 16.63 & 17.82 & 17.75 & 31.83
				\\	
				HCML~\cite{tone} & 21.70 & 22.24 & 24.44 & 20.08
				\\
				BDTR~\cite{VT} &32.72 & 31.10  & 33.47 & 31.83
				\\	
				D$^2$RL~\cite{GANCVPR19} &  - & - & 43.4 & 44.1
				\\
				DGD~\cite{DGD}  &  - & -& 31.94&35.65
				\\
				MSR~\cite{msr2019tip}  &  - & - &  48.43 & 48.67
				\\
                Triplet Loss &  50.72 &  50.55 &  57.95 & 57.70
                \\
                BNNeck~\cite{luohao} & 46.50 &46.97  & 57.46 & 58.54
                \\
				AT (\textbf{Ours})& \underline{65.65} & \underline{65.39} & \textbf{69.60} & \textbf{69.84}
				\\
				expAT (\textbf{Ours})& \textbf{67.45} & \textbf{66.51} & \underline{66.48} & \underline{67.31}
				\\
				\toprule
	\end{tabular}}}
	\label{tab_q2_regdb}
	\vspace{-1.1em}
\end{table}

\begin{table}[h]
	\centering
	\caption{Peer comparison on the Market-1501 dataset and DukeMTMC-reID dataset.  Rank-1 accuracy~(\%) and  mAP (\%) are listed. ``CL'' means using Center Loss~\cite{centerloss}.  The best results are highlighted in bold  and the second best are underlined.}
	\scalebox{1}{
		\setlength{\tabcolsep}{3mm}
		{
			\begin{tabular}{c|cc|cc}
				\bottomrule
				\multirow{2}{*}{Method}&\multicolumn{2}{c|}{Market-1501} &\multicolumn{2}{c}{DukeMTMC-reID}
				\\
				\cline{2-5}
				& Rank-1   & mAP & Rank-1   & mAP
				\\
				\hline
				PCB-RPP~\cite{pcbrpp2018} & 93.8 & 81.6 & 83.3 & 69.2
				\\	
				MGN~\cite{MGN} & \textbf{95.7} & \textbf{86.9}  & \textbf{88.7} & \textbf{78.4}
				\\
				HA-CNN~\cite{li2018harmonious} & 91.2 & 75.7 & 80.5 & 63.8
				\\
				BDB~\cite{BDB} & 94.2 & 84.3 & 86.8 & 72.1
				\\			
				Cam-GAN~\cite{camgan} & 88.1 & 68.7 & 75.3 & 53.5
				\\
				LSRO~\cite{zheng2017unlabeled} & 84.0 & 66.1 & 67.7 & 47.1
				\\
				\hline
				triplet baseline~\cite{luohao} & 92.0   & 81.7 & 82.6 & 70.6
				\\
				BNNeck~\cite{luohao} & 94.1 & 85.7 & 86.2 & 75.9
				\\
				\hline	
				AT & 94.5 & 86.3 & 86.7 & 76.5
				\\
                expAT\_alone & 91.20 & 80.40 & 82.9 & 70.4
                \\
                expAT\_l2norm  & 62.10& 44.30 & 52.8 & 33.4
                \\
				expAT (\textbf{Ours})& \underline{94.7} & 86.6 & \underline{87.6} & \underline{77.1}
				\\
				\hline
			    BNNeck+CL~\cite{luohao} & 94.5 & 85.9 & 86.4 & 76.4
				\\
				AT+CL  & 94.6 & \textbf{86.9} & 86.8 & 76.5
				\\
				expAT+CL (\textbf{Ours}) & \underline{94.7} & \underline{86.8} & 87.4 & 76.9
				\\
				\toprule
	\end{tabular}}}
	\label{tab_q2_market}
	\vspace{-1.1em}
\end{table}

The quantitative results are shown in Table~\ref{tab_q2}. As most hand-crafted feature based methods are specifically designed for the RGB domain taking advantage of color information, which is unavailable in the infrared modality, they show humble performance on the RGB-IR Re-ID task. On the contrary, as the deep learning based methods are data-driven and able to learn geometric and semantic  features adaptively for different modalities, they achieve much stronger performance.

The proposed methods (AT and expAT) outperform all the state-of-the-art single-stream methods by a large margin, and achieve competitive results compared with MSR~\cite{msr2019tip}.
Notably, MSR adopts two special backbone networks~\cite{DGD} and decouples feature from different modalities, while our method is implemented on a single-stream ResNet-50 to process data from both modalities.
Although GAN based method~\cite{cmgan} shows fair performance compared with other methods, they take long time to converge (cmGAN takes more than 2000 epochs to converge as stated in the original paper).
Correspondingly, the proposed method takes less than 13 epochs to converge, and increases the  rank-1 accuracy and mAP dramatically. With these statistics results, we conclude that expAT loss shows the state-of-the-art performance on the cross-modality RGB-IR Re-ID task.

\begin{figure}[t]
	\centering
	\includegraphics[scale=0.33]{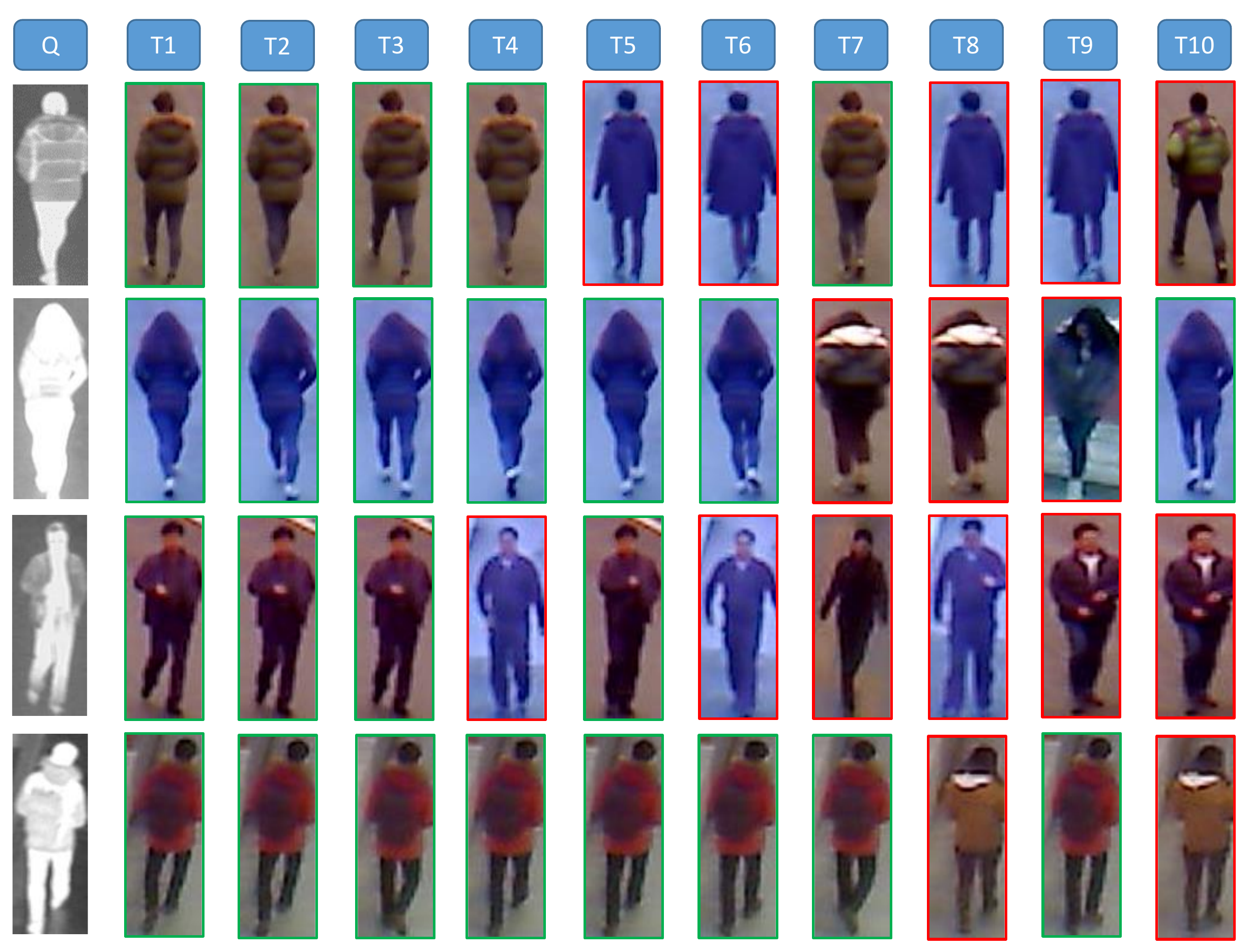}
	\caption{Retrieval results of expAT Loss on RegDB dataset under Thermal to RGB setting.
    In each row, top-\emph{k} images are arranged in ascending order according to their distances with the query on the left.
    The true and false matches are in the green and red boxes, respectively.}
	\label{img_res_vis_regdb}
\vspace{-1.em}
\end{figure}

\subsubsection{Comparison on RGB-thermal Re-ID}
To verify the effectiveness of our method, we evaluate it on the RegDB dataset~\cite{nguyen2017} in Table~\ref{tab_q2_regdb} and provide some retrieval results in Figure~\ref{img_res_vis_regdb}.
The proposed methods yield the best results with significant margin compared with other studied algorithms.
Compared with bi-directional Triplet Loss, the proposed expAT Loss improves the Rank-1 / mAP from 50.72\% / 50.55\% to 67.45\% / 66.51\% using exactly the same backbone model and training practices under ``Thermal to RGB'' setting.

As shown in Figure~\ref{img_res_vis_regdb}, thermal images contain little color information, while the contour information is vague.
The model makes mistakes when different identities wear similar clothing (e.g. T5 in the first row).
The difference in the imaging principles of RGB cameras and thermal cameras also leads to distinct appearances of the same person.
Take the third row of the figure as example, wrong matching (T9) even looks more like the query than correct matching (T1).
This is because the RGB camera does not capture the clothing pattern at the belly clear enough, while the thermal camera has higher response at this area because of the body  temperature.

\subsubsection{Comparison on RGB Re-ID}
The criterion of mining an angularly discriminative common feature space is also applicable for the single-modality Re-ID task.
Thus, AT loss and expAT loss can be generalized to the task of single-modality person re-identification. To further examine the validity of our methods, we implement them on a Resnet-50 backbone following the setting in BNNeck~\cite{luohao}, and compare with the latest Re-ID algorithms on two popular benchmarks (Market-1501~\cite{market} and DukeMTMC-reID~\cite{zheng2017unlabeled}). The compared methods include the part based (PCB-RPP~\cite{pcbrpp2018}, MGN~\cite{MGN}), attention based (BDB~\cite{BDB}, HA-CNN~\cite{li2018harmonious}) and GAN based (Cam-GAN~\cite{camgan}, LSRO~\cite{zheng2017unlabeled})  methods.

As shown in the Table~\ref{tab_q2_market}, the proposed methods yield very strong performance and is compatible another effective loss function (Center Loss~\cite{centerloss}).
Compared with triplet loss baseline, expAT loss improves the rank-1 accuracy and mAP from 92.0\% and 81.7\% to 94.7\% (+\textbf{2.7}\%) and 86.6\% (+\textbf{4.9}\%) on the Market-1501 dataset. On the DukeMTMC-reID dataset, rank-1 accuracy and mAP increase from 82.6\% and 70.6\% to 87.6\% (+\textbf{5.0}\%) and 77.1\%(+\textbf{6.5}\%). Although MGN~\cite{MGN} achieves the best results,  it is worth to point out that MGN adopts a much more complicated framework with 8 branches. Our approach shows competitive performance with a single-stream Resnet-50 backbone. These results indicate that expAT loss is highly effective for the single-modality RGB Re-ID task.

It is also noteworthy that, for ablation study, using only expAT Loss wihout CSBN (expAT\_alone) or using L2 normalization instead of CSBN (expAT\_l2norm) achieves much worse performance.
This phenomenon supports our study on the cross-modality task in Table~\ref{tab_q1}.

\section{Conclusions}
This work focuses on a challenging newly-developing task: RGB-IR Re-ID.  In this paper, the Bi-directional Exponential Angular Triplet~(expAT) Loss is proposed to help deep networks learn angularly representative embedding vectors of images from different modalities. The expAT loss directly constrains the included angles between embedding vectors and helps partition the feature space angularly.
Moreover, a common space batch normalization is adopted to help expAT loss stabilize and learn the magnitudes of embedding vectors. Extensive experiments and feature visualization are performed, which validate the effectiveness of the proposed methods.

\bibliographystyle{IEEEtran}
\bibliography{atloss}

\begin{IEEEbiography}[{\includegraphics[width=1in,height=1.25in,clip,keepaspectratio]{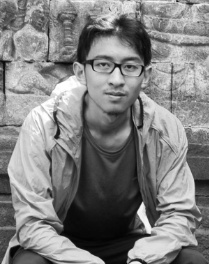}}]{Hanrong Ye} received the B.S. degree from School of Physics, Sun Yat-sen University (SYSU), China, with iSEE Lab. 	He is a research graduate student studying at Peking University (PKU), China.
 His research interest lies in machine learning and computer vision.	
\end{IEEEbiography}

\begin{IEEEbiography}[{\includegraphics[width=1in,height=1.25in,clip,keepaspectratio]{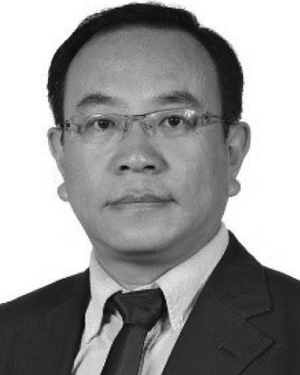}}]{Hong Liu} (M'08) received the Ph.D. degree in mechanical electronics and automation in 1996.
	He is currently a Full Professor in the School of Electronics Engineering and Computer Science, Peking University (PKU), Beijing, China. He has been selected as Chinese Innovation Leading Talent supported by ``National High-level Talents Special Support Plan'' since 2013. He is also the Director of Open Lab on Human Robot Interaction, PKU. He has published more than 150 papers. His research interests include computer vision and robotics, image processing, and pattern recognition. He received the Chinese National Aero-space Award, the Wu Wenjun Award on Artificial Intelligence, the Excellence Teaching Award, and the Candidates of Top Ten Outstanding Professors in PKU. He is the Vice President of Chinese Association for Artificial Intelligent (CAAI), and the Vice Chair of Intelligent Robotics Society of CAAI. He has served as keynote speakers, Co-Chairs, Session Chairs, or PC members of many important international conferences, such as IEEE/RSJ IROS, IEEE ROBIO, IEEE SMC, and IIHMSP, and serves as reviewers for many international journals such as Pattern Recognition, the IEEE Transactions on Signal Processing, and the IEEE Transactions on Pattern Analysis and Machine Intelligence.
\end{IEEEbiography}

\begin{IEEEbiography}[{\includegraphics[width=1in,height=1.25in,clip,keepaspectratio]{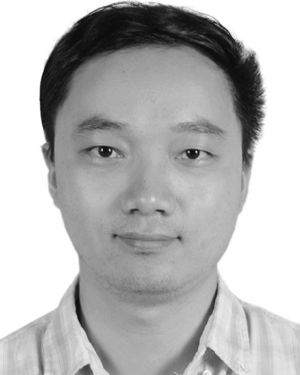}}]{Fanyang Meng} received the Ph.D. degree from the School of EE\&CS, Shenzhen University, China,	in 2016, under the supervision of Prof. X. Li.
	He currently serves as a assistant research fellow in Peng Cheng Laboratory. His research interests include video coding, computer vision, human action recognition, and abnormal detection using RGB, depth, and skeleton data.
	
	His research interests include computer vision, human action recognition, and abnormal detection using RGB, depth, and skeleton data.
\end{IEEEbiography}

\begin{IEEEbiography}[{\includegraphics[width=1in,height=1.25in,clip,keepaspectratio]{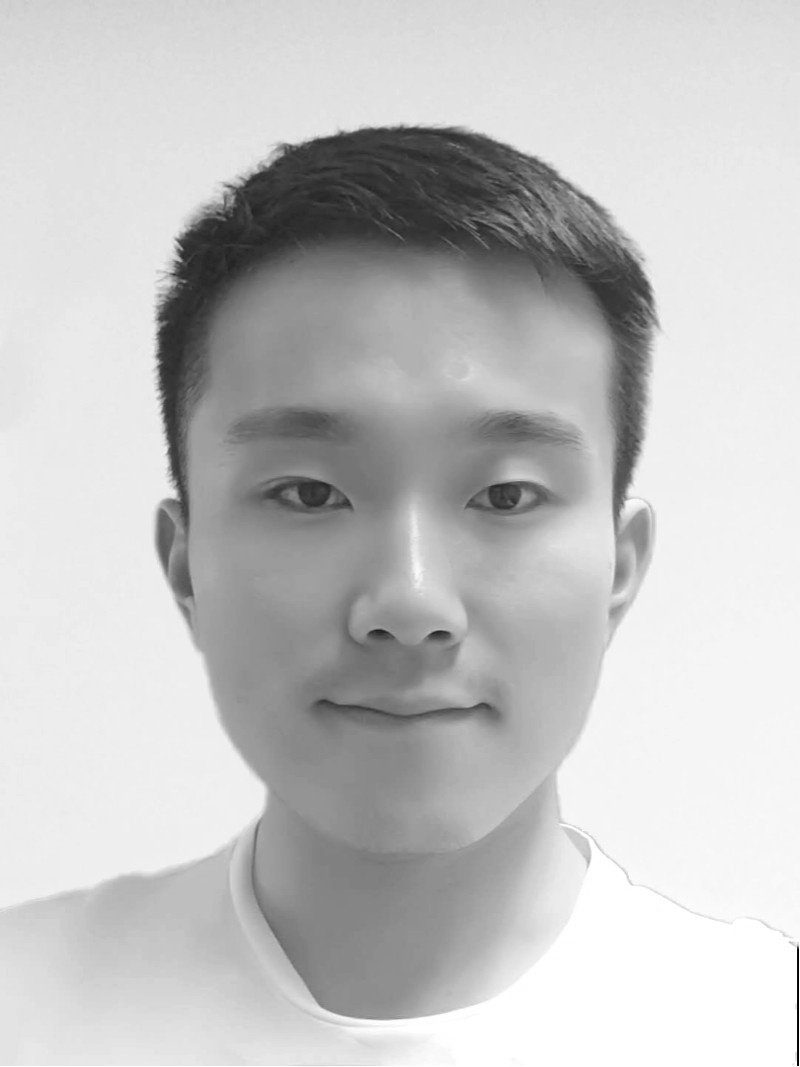}}]{Xia Li}  received his bachelor's degree from Beijing University of Posts and Telecommunications (BUPT) in 2017. He is currently pursuing his master's degree  at Peking University (PKU), China, advised by  Prof. Hong Liu and Prof. Zhouchen Lin.
	
His research interest lies in Computer Vision, Semantic Segmentation and Low-level Vision. He has published papers on academic conferences including ECCV' 20, CVPR' 20, AAAI'20, ICCV'19, MICCAI'19 and ECCV'18.
	
\end{IEEEbiography}

\end{document}